\documentclass{article}
\usepackage{graphicx}
\usepackage{amsmath}
\usepackage{amsfonts}
\usepackage{amssymb}
\usepackage{amsthm}
\usepackage{mathtools}
\usepackage{enumerate}
\usepackage{lscape}
\usepackage{longtable}
\usepackage{rotating}
\usepackage{multirow}
\usepackage{color}
\usepackage{url}
\usepackage{subfigure}
\usepackage{rotating}
\usepackage{comment}
\usepackage{graphicx}
\usepackage{tikz}
\usetikzlibrary{arrows,positioning}
\usepackage[utf8]{inputenc}

\usepackage{xcolor}

\title{``Cause'' is Mechanistic Narrative within Scientific Domains: An Ordinary Language Philosophical Critique of ``Causal Machine Learning''}
\author{Vyacheslav Kungurtsev \footnote{VK, GS, and MK are at the Department of Computer Science, Czech Technical University in Prague, in Czechia. LCM is affiliated with the Institute of Advanced Consciousness Studies in Los Angeles, California, USA}, Leonardo Christov Moore, \\ Gustav {\v S}{\' i}r, and Martin Krutsk{\' y}}
\begin{document}

\newcommand{\todolm}[2][]
{\todo[color=green!25, #1]{Leo: #2}}
\newcommand{\todosl}[2][]
{\todo[color=red!25, #1]{Slava: #2}}

\maketitle

\begin{abstract}
    Causal Learning has emerged as a major theme of research in statistics and machine learning in recent years, promising computational techniques to reveal ``true'' causality. In this paper, we critique the premise of causal learning by considering the epistemology of causality across disciplines, applying the Ordinary Language method of an anthropological investigation of customary word use to investigate valid semantics of reasoning about cause and effect in the real world. 
We observe that although cause-and-effect semantics vary between scientific domains, they maintain a consistent central function of describing the mechanisms underlying forces most salient to the systems the scientific domain studies.
    A critical distinction is the degree to which the mathematical representations of the models used for statistical analysis exhibit a faithful correspondence to expert knowledge in mechanism. We demarcate 1) physics and engineering as domains wherein mathematical models are sufficient to comprehensively describe causality, in contrast to 2) biology, which studies open and irreducible systems with mechanisms crossing scales through emergence, and 3) the social sciences as suffering from compounding difficulties for precision but providing, through Hermeneutics, the potential for subjective phenomenology yielding findings instrumentally useful when interpreted through individual thought schemas and compelling narratives. We posit the greater the discrepancy between expert-defined models and complete characterization of mechanism, the more that epistemic virtue requires that definitive causal claims regarding  phenomena can only come through an agglomeration of consistent evidence across multiple domains. 
    Exercising greater caution in communicating the degree of certainty evidence provides, especially demanding restraint in the face of incentives to overstate research conclusions, is the only durable solution to modern science's collective action problems. 
\end{abstract}

\section{Introduction}
 A recent popular interest in Artificial Intelligence (AI) and Machine Learning (ML) has been the attempt to find \emph{causal} relationships, rather than ``mere'' statistical association. While this may seem reasonable to a lay person, singly defining causality is not straightforward. Still, dominant trends in the statistical literature on causalitybased on independence criteria, have been touted as discovering causal phenomena and thus improving the foundational understanding fields ranging from physics to biology and economics ~\cite{scholkopf2021toward}. 

The independence criteria that lie at the foundation of causal inference and discovery ~\cite{pearl2009causality} and~\cite{peters2017elements}, entail the axiomatic assertion that a statistically significant model wherein there is a graph defining influence, where certain conditional independence criteria hold, performs a certified identification of a cause and effect dynamic between two measured (or latent) variables. 

However, contemporary causal inference has known fundamental issues. First, in practice, it is very rare for a paper performing causal learning on real data to identify a graph with all conditional independence hypothesis tests passing (the authors were not able to identify one). Next, the fundamental assumption of causal sufficiency does not hold for most phenomena of interest. 

The rich discussions of causality in Epistemology and the philosophy of science make clear that a ``cause'' is a deeper understanding of a system than mere observation of phenomena appearing together, and discussions of how one can know something causes something else are prominent in the philosophical canon, and differently underly the domains of human inquiry. 

In this paper, we attempt to elucidate the semantics and function of "causality". In particular, after reviewing the formal definitions of causal inference and learning in contemporary causal machine learning, we proceed with a review of significant perspectives of causality through the history of (mostly western) philosophy, and their manifestation in scientific disciplines. 

We shall present a claim in the spirit of \emph{ordinary language philosophy}(REF) on the necessity of a pluralistic definition of causality inseparable from the domain of inquiry. 
One can speak of cause and effect in a variety of domains, from ordinary lay physics (``the wind caused it to fall'') and psychology (``his anger caused conflict'') to scientific physics, chemistry, ecology, economics, and sociology. There are enough significant distinctions in what is meant by cause, e.g., decay in particle physics, or increase of mortality rates due to economic turbulence, to warrant distinct quantitative tools of discovery and inference. Yet, despitedistinctions due to semantic diversity in scientific grammar, the function of narratives of cause and effect appear to have conformity in the abstract. Namely, causes are the central mechanisms by which external inputs or internal dynamics can trigger either instant changes to direction and/or magnitude of the dynamic evolution of a system and its components. However, the utility of a causal description within a scientific tribe, clinical practitioner of the material studied in the science, cross-disciplinary interaction, and public communication, is a necessary consideration for notions of cause and effect to have any practical sense.


We will also cover insights from what could be loosely denoted as anti-realist interpretations of science by Thomas Kuhn and Epistemology by Willard Quine. Here, one cannot perform induction towards causality without significant inductive bias, corresponding in practice to model representation in technical induction. The model representation, e.g. differential equations, is meant to correspond to the structure of causal semantics within the scientific domain. Aligning empirical observation to actually learn about the world requires some methodological choice of statistical model representation.

We can see the ``causal learning'' graphical statistical independence criteria as one such choice of inductive bias that is fundamentally axiomatic, and not a choice of any practice-independent certification of its veracity. Moreover, the farther that directed graphs with Markov transitions are from the canonical functional equation systems of any scientific domain, the more inappropriate and overconfident causal learning of this form becomes.


We then review primary scientific domains and their semantics of causality. We highlight the uniqueness of Physics as defining equations that encompass the entire phenomenon of causality in the system, where Biology, in contrast, involves open systems and multiple scales of phenomena with unidentified influences of emergence. 

Thus the Epistemological strength of the a priori inductive bias of the system as defined by the appropriate scientific paradigm can be ascertained in one of two ways:  1) whether it is a closed form (free from interaction with an external environment) system, and 2) whether a well-defined mathematical structure is available. When these circumstances are less favorable for definitive conclusions on cause and effect, as in Medicine, we observe how identifying consistent influences at multiple scales has been central to establishing sound conclusions regarding causes.

The same can be observed in the social sciences, currently suffering from a replication crisis. Since the complexity of a human being makes faithful modeling of the system impossible, no statistical finding or theoretical model can establish strong evidence for any cause-effect conjecture in the field. The accumulation of consistent findings across multiple scales and scientific grammars is thus necessary to establish generic facts about social and psychological phenomena. Fortunately, a liminal category of language games, Hermeneutics, presents the possibility of domain-situated inquiry as providing useful application to individuals for their decisions, even without an established, precise ontology. Only protracted multi-disciplinary research work with quantitative methods together with historical descriptions, theoretical models, associated physio-biological mechanisms, and neurophenomenology, can, through multi-domain harmony, attain reasonable confidence and certainty regarding social science conjecture. We present cognitive science as a crystallizing contemporary example of a comprehensively multi-disciplinary field in this vein, and review the varied semantics of causation appearing across its subspecialties. 

Last, we present some considerations as far as how cause can be related across disciplines, and the role of classical causal inference and other modern statistical techniques. We note that the methods of causal inference are still useful and powerful statistical methods, they are simply too overstated to consider to certify a causal phenomenon. Through careful use of language, and proper respect to the manner in which terminology is used across the semantics of the various domains of scientific experts, we can take proper caution to ensure intellectual discussion of the results of the scientific and machine learning literature is in proper harmony.

Importantly, these investigations into multidomain causality languages present new approaches for explainability -- the field of AI concerned with endowing fitted models to data with understandable language or GUI output -- as a useful and important field of AI for analysis and methodological development. Using techniques in neuro-symbolic modeling~\cite{besold2017reasoning}, one can use logics to define and integrate knowledge extracted from narrative descriptions and highly parameterized empirical models associated with particular domains.

\section{Background}
\subsection{Causal Machine Learning}

``Causal inference'' and ``causal discovery'' in machine learning describe a paradigm of inductive research that has recently risen in prominence, endorsement from leaders in the field~\cite{scholkopf2021toward} and institutional acceptance for critical decision making, e.g. in medicine and public policy \cite{lechner2023causal,sanchez2022causal}. Its axioms are largely informed by the framework developed by Pearl~\cite{pearl2009causality}, which uses \textit{Probabilistic Graphical Models (PGMs)}, specifically \textit{Bayesian Networks (BNs)}, to represent and reason about causal relationships, modeled as a set of probabilistic dependencies among variables, structured in a graphical form.

\paragraph{Bayesian Networks (BNs)} are a class of models that represent a set of random variables and their conditional dependencies. Each node in a Bayesian Network corresponds to a random variable, while directed edges (arrows) between nodes signify probabilistic dependencies. This structure encapsulates the joint probability distribution of the variables involved, enabling efficient reasoning about uncertainty in conditional and marginalized probabilistic relationships. Mathematically, the joint probability distribution of a set of variables in a BN can be factorized as:

\[
P(X_1, X_2, \ldots, X_n) = \prod_{i=1}^{n} P(X_i \mid \text{Pa}(X_i)),
\]

\noindent where \( \text{Pa}(X_i) \) denotes the set of parents (direct predecessors) of node \( X_i \). This factorization enables simpler computation by leveraging the conditional \textit{independencies} inherent in the structure of the network.

\textit{Conditional Independence} is a key concept that allows the simplification in the joint probability distribution factorization.
It is closely related to the concept of \textit{D-separation}, a graphical criterion for determining whether two sets of variables are independent given a third set. If all paths between two variables \( X \) and \( Y \) are blocked by conditioning on a third set \( Z \), then \( X \) and \( Y \) are conditionally independent, denoted as \( X \perp Y \mid Z \). This graphical criterion is central for defining the causal structure of a system by this schema.

In the context of causality, one can reason about the \textit{Statistical Significance} of the correspondence of the BN to some causal model of the underlying system. This can be formally defined as a DAG wherein \textit{each} edge represents a causal dependency between the random variables (RVs), validated through statistically significant conditional independence tests.
Each edge \( X_i \to X_j \) in the graph \( G \) must satisfy the conditional independence criterion:
\[
X_j \not\perp X_i \mid \text{Pa}(X_j) \setminus \{X_i\},
\]
which is validated using statistical hypothesis tests (e.g., chi-square tests, partial correlation, or kernel-based independence measures). These tests confirm that the observed data supports the conditional dependencies implied by the DAG. Importantly, \( G \) must align with the data such that no implied conditional independence relations are contradicted by the empirical evidence.

A key assumption underlying the construction of such causal models is \textit{Causal Sufficiency}, which requires that \textit{all} common causes (latent variables) of the observed variables are included in the observed data.
Formally, let \( X \) and \( Y \) be two observed random variables, and let \( Z \notin \{X, Y\} \) be a {latent} variable that causally influences both \( X \) and \( Y \). The assumption of causal sufficiency implies that:
\[
X \not\perp Y \mid Z, \quad \forall Z,
\]
where \( Z \) comprises \textit{all} common causes of \( X \) and \( Y \), such that no unmeasured common cause exists for any of the observed variables. In practical terms, this assumption ensures that the DAG \( G \) accurately captures {all} direct and indirect causal relationships without hidden confounds. While unrealistic, this assumption is important because unmeasured confounds can create spurious associations between variables, leading to biased causal estimates.\footnote{As we will discuss, causal sufficiency rarely holds in practice since many systems of interest are \textit{open systems}, where latent factors interact with the observed variables. For example, in social sciences or medicine, environmental influences, genetic factors, or historical contingencies serve as latent confounds that are impractical to measure or include in the analysis. That the violation of this formal assumption is endemic in most scientific domains hints at the limitation of this axiomatic approach to causation.}

\textit{Dynamic Bayesian Networks (DBNs)} extend the Bayesian Network framework to model \textit{sequences} of random variables, i.e. parameterized stochastic processes, making them particularly useful for temporal data. In DBNs, the network is divided into time slices, with the dependencies between variables at one time step influencing those at the next. A DBN consists of both \textit{static random variables}, which do not change over time, and \textit{time-dependent random variables}, which evolve over time. The joint distribution of a DBN can be expressed as:

\[
P(X_{1:T}) = P(X_1) \prod_{t=1}^{T-1} P(X_{t+1} \mid X_t, \text{Pa}(X_{t+1})),
\]

where \( X_{1:T} \) represents the sequence of variables from time \( 1 \) to \( T \), and each \( X_{t+1} \) depends on a set of parent nodes, that includes some subset of variables at the current time $t+1$, \emph{instantaneous causes}, another subset of variables evaluated at the previous time $t$ as well as lagged (including auto-regressive) effects $t'<t$, and, potentially, static (non time-varying) variables. For instance, the progression of a disease $X^{(D)}_{t}$ to $X^{(D)}_{t+1}$ depends on the current state of the disease $X^{(D)}_{t}$, as well as momentum in its progression $X^{(D)}_{t}-X^{(D)}_{t-1}$, the robustness of the current immunity resistance, together with the current and past concentration of T-cells released to combat disease pathogens.

Stationary and non-stationary Dynamic Bayesian Networks are distinguished according to whether the structure and magnitude of the graphical links between variables changes over time. Learning stationary networks is far more tractable and effective, as data over long time horizons presents an appreciable quantity of independent samples for estimating the transition probability function. 


A notable tool in causal inference is Pearl’s \textit{"do" calculus}, which formalizes the effect of \textit{interventions} in a system. In this context, the notation \( do(X=x) \) represents an intervention where variable \( X \) is forcibly set to value \( x \). The distinction between observational data and experimental data is captured by the equation:

\[
P(Y \mid do(X=x)) \neq P(Y \mid X=x),
\]
which emphasizes that interventions can alter the distribution of the system in ways that simple observation cannot.

\paragraph{Example of causal inference with a DBN}

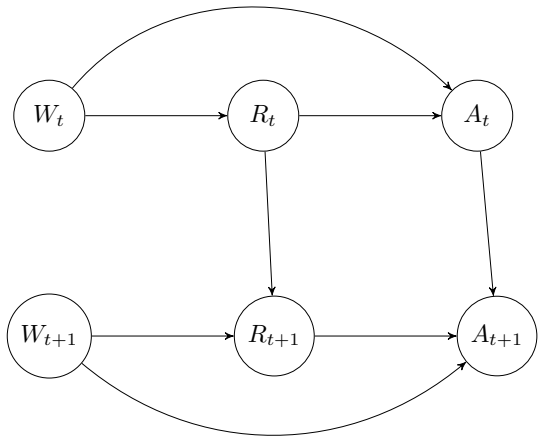
\begin{figure}[h!]
\centering
\resizebox{0.6\textwidth}{!}{
\begin{tikzpicture}[
    ->, >=stealth', auto, node distance=2cm, every node/.style={circle, draw, minimum size=1cm}]
    \node (Wt) {$W_t$};
    \node (Rt) [right=of Wt] {$R_t$};
    \node (At) [right=of Rt] {$A_t$};
    \node (Wt1) [below=of Wt] {$W_{t+1}$};
    \node (Rt1) [right=of Wt1] {$R_{t+1}$};
    \node (At1) [right=of Rt1] {$A_{t+1}$};
    \path[->] (Wt) edge (Rt);
    \path[->] (Wt) edge[out=50] (At);
    \path[->] (Rt) edge (At);
    \path[->] (Rt) edge (Rt1);
    \path[->] (At) edge (At1);
    \path[->] (Wt1) edge (Rt1);
    \path[->] (Rt1) edge (At1);
    \path[->] (Wt1) edge[out=-40,in=220] (At1);
\end{tikzpicture}
}
\caption{A simple DBN model of dependencies over two consecutive timesteps.}
\label{fig:DBN}
\end{figure}

Let us illustrate a simple causal system using a DBN. At each time slice \( t \), the system includes three variables: \( W_t \) (Weather), \( R_t \) (Road Conditions), and \( A_t \) (Traffic Accidents). Weather \( W_t \) directly influences Road Conditions \( R_t \), as rainy weather can cause slippery roads. Additionally, \( W_t \) independently affects \( A_t \), representing the direct impact of weather on accidents (e.g., reduced visibility). Finally, \( R_t \) influences \( A_t \), modeling the effect of slippery roads on accidents. Temporal dependencies connect the states across time: \( R_t \) affects \( R_{t+1} \), and \( A_t \) affects \( A_{t+1} \).
The joint probability distribution for two time slices (\( t \) and \( t+1 \)) is then:
\begin{align*}
P(W_t, R_t, A_t, W_{t+1}, R_{t+1}, A_{t+1}) = P(W_t) P(R_t \mid W_t) P(A_t \mid W_t, R_t) \\
P(W_{t+1}) P(R_{t+1} \mid R_t, W_{t+1}) P(A_{t+1} \mid W_{t+1}, R_{t+1}).
\end{align*}
with the DBN structure illustrated in Figure~\ref{fig:DBN}.

Suppose now that we intervene to make the roads ``Normal'' (\( R_t = \text{Normal} \)), irrespective of the weather. This breaks the indirect causal link between \( W_t \) and \( R_t \), isolating the direct impact of weather on accidents:
\[
P(A_t \mid do(R_t = \text{Normal})) = P(A_t \mid W_t, R_t = \text{Normal}).
\]

\paragraph{Dynamic Systems Perspective}

The definitions of Bayesian Networks (BNs) and Dynamic Bayesian Networks (DBNs) introduced above can also be interpreted within the framework of Structural Equation Models (SEMs), also referred to as \textit{Structural Causal Models} (SCMs) in the context of causality~\cite{peters2017elements}.
These models generalize BNs~\cite{bongers2021foundations} by explicitly introducing structural equations to describe the relationships between variables. This can be used to define and model \textit{dynamic systems}, which evolve over time through deterministic or probabilistic mechanisms.
An SCM generally consists of a set of structural assignments that define each variable as a deterministic function of its direct causes and an independent \textit{noise} term. For dynamic systems, SCMs then extend to describe temporal relationships between variables across time steps. 

In dynamic systems, the state of each variable \( X_t \) at time \( t \) is determined by a structural equation of the form:
\[
X_t = f(X_{t-1}, U_t),
\]
where \( X_{t-1} \) represents the system's state at the previous time step, \( U_t \) denotes independent noise terms, and \( f \) is the causal mechanism defining the evolution of the system. This formulation naturally extends the graphical representation of DBNs by encoding temporal dependencies through directed edges across time slices and ensuring that the joint distribution of the system still factorizes as:
\[
P(X_{1:T}) = P(X_1) \prod_{t=1}^{T-1} P(X_{t+1} \mid X_t, \text{Pa}(X_{t+1})).
\]
For systems operating in continuous time, the evolution of variables can then be expressed using Ordinary Differential Equations (ODEs). In this framework, the dynamic behavior of a variable \( X(t) \) is described by:
\[
\frac{dX(t)}{dt} = g(X(t), U(t)),
\]
where \( g \) defines the rate of change of \( X(t) \) and \( U(t) \) captures external noise or exogenous inputs. This ODE-based representation provides a natural extension of discrete-time SCMs, where the discrete equations can be viewed as time-discretized approximations of the continuous-time dynamics. Importantly, this approach enables modeling of complex systems with smooth trajectories, such as those encountered in physics, biology, and other fields where time-continuous processes dominate.

Building on the structural formalism of SCMs, \textit{Additive Noise Models} (ANMs) provide a contemporary framework for \textit{causal inference} in such systems, particularly effective in nonlinear settings~\cite{hoyer2008nonlinear}. It builds on the idea that correlation and causation can be told apart as the causal direction can be inferred by leveraging the independence of noise in additive models. Particularly, an ANM assumes that for a given causal relationship:
\[
Y = f(X, U) = f(X) + U,
\]
where \( U \) is an independent noise term, it holds that \( U \perp X \). This independence introduces an asymmetry that helps to identify the causal direction. If \( Y \to X \), then reversing the direction typically violates the independence of \( U \), as:

\[
P(X \mid Y) \neq P(X \mid f(Y) + U).
\]
In practical applications, residual independence tests, which assess whether \( U \) remains independent of \( X \), are employed to confirm the causal direction.
For dynamic systems, the ANM framework extends naturally to time-evolving relationships. In a DBN, temporal dependencies can be modeled as:
\[
X_{t+1} = f(X_t, \text{Pa}(X_t)) + U_t,
\]
where \( \text{Pa}(X_t) \) includes static and temporal parent nodes, and again \( U_t \perp X_t \). Combined with assumptions on temporal smoothness and structural sparsity, ANMs can typically improve causal inference in time-series data.

This perspective of learning representation structure and parameters of a set of ODEs describing a system is central to contemporary causal machine learning for scientific domains, see e.g.~\cite{yao2024marrying,pervez2024mechanistic}. The correspondence in Probabilistic Graphical Models and Structural Equation Models indicates their potential for intuitive interpretability. However, we shall see that the dynamic systems perspective is ultimately an analogy taken unscrupulously from physics, and the semantics through which it aligns physical intuition and fundamental theory with dynamic evolution systems by a narrative of causation. The structural features and foundational assumptions of this correspondence are unique to physics, and are less certified in their presence, or even known to be false, in other domains. 

This is associated with another important consideration of SEM/BN modeling -- \textit{sparsity}. In defining a graphical model to represent random variables, the presence or absence of edges between quantities defines the primary mechanisms of influence in the dynamics of the quantities of interest. Thus the task of sparse representation learning, inspired by the classical field in signal processing of compressed sensing, is a pillar of inductive causal inference. The greater interpretability, or ease of conceptually understanding, of sparse models presents the natural task  of ``disentangling'' the complex relationships between variables as instrumental to understanding the essential processes of the system~\cite{van2019disentangled}. This hints the fundamental dissatisfaction with such metrics for ascertaining cause and effect, they intuitively seem related, and as notable characterizations of well-respected scientific explanations, even identifying or at least encouraging features of ``good'' causal narratives, however it is clear that parsimony is not the definition of cause and effect, and is not \emph{the} defining feature.

\paragraph{Causal Probabilistic Logic}
Recall that a (D)BN defines a causal model through its nodes and edges. \emph{Lifting} constitutes studying abstractions of the resulting model corresponding to patterns in the graphical structure of causal links that are repeated within the entire graph, oftentimes in association with certain values of other graph nodes~\cite{kimmig2015lifted}. Lifting (D)BNs presents the circumstances for the development of general rules and formulae that concern causal relationships between abstractions within the domain. Specifically, \textit{Causal Probabilistic Logic}~\cite{vennekens2009cp} (CP-logic) offers a logical formalism for representing probabilistic causal laws. Developed as a compact and flexible representation of dynamic probabilistic systems, CP-logic extends the language of \textit{logic programming}~\cite{doets1994logic} to explicitly capture causal relationships, including their probabilistic nature. This provides a bridge between the common paradigm of \textit{probabilistic logic programming}~\cite{de2015probabilistic} and causal inference.

Syntactically, in CP-logic, a probabilistic causal law is represented as a statement of the form:
\[
\forall x \; (A_1 : \alpha_1) \lor (A_2 : \alpha_2) \lor \dots \lor (A_n : \alpha_n) \leftarrow \phi,
\]
where  \(x\) is a tuple of all the, universally quantified, variables within the law; \(\phi\) is a first-order (lifted) formula representing the preconditions (cause) of the event; \(A_i\) are the logical atoms representing potential effects of the event; and \(\alpha_i \in [0, 1]\) are the probabilities associated with each effect \(A_i\), \(\sum_{i=1}^n \alpha_i \leq 1\).
This statement reads as: ``For all \(x\), if the condition \(\phi\) holds, then an event occurs with one of the effects \(A_1, A_2, \dots, A_n\), each with probability \(\alpha_i\).''

Semantically, CP-logic defines causal probabilistic laws in terms of \emph{probability trees}, which represent the evolution of states in a dynamic system. Note that, using the probability trees as the basic semantic objects, this view differs from Pearl's approach through Bayesian networks where causal information is viewed more statically, or at least as a stochastic process with stationary transitions, with probability distributions as the basic semantic objects.
Each node in the tree corresponds to a state, and transitions between nodes are driven by causal events governed by the CP-laws. The root of the tree represents the initial state, while subsequent nodes reflect transitions based on probabilistic outcomes. The probability of reaching a specific leaf node can then be computed as the product of probabilities along its path.

\begin{figure}[h!]
\centering
\resizebox{0.8\textwidth}{!}{
\begin{tikzpicture}[
  level distance=2.5cm,
  level 1/.style={sibling distance=6cm},
  level 2/.style={sibling distance=3cm},
  edge from parent/.style={draw, -latex},
  every node/.style={minimum width=2cm, minimum height=1cm, align=center}
  ]

\node (Start) {Root}
  child {node (Throws) {\texttt{Throws(Mary)}}
    child {node (Break) {\texttt{Break}}
      child [grow=down, level distance=1.5cm] {node (BreakP) [draw=none] {\(P = 0.4\)}}
    edge from parent node[midway, left] {\(P = 0.8\)}}
    child {node (NotBreak) {\(\neg\)\texttt{Break}}
      child [grow=down, level distance=1.5cm] {node (NotBreakP) [draw=none] {\(P = 0.1\)}}
    edge from parent node[midway, right] {\(P = 0.2\)}}
  edge from parent node[midway, left] {\(P = 0.5\)}}
  child {node (NotThrows) {\(\neg\)\texttt{Throws(Mary)}}
    child {node (NotBreakAlt) {\(\neg\)\texttt{Break}}
      child [grow=down, level distance=1.5cm] {node (NotBreakAltP) [draw=none] {\(P = 0.5\)}}
    edge from parent node[midway, right] {\(P = 1.0\)}}
  edge from parent node[midway, right] {\(P = 0.5\)}};

\end{tikzpicture}
}
\caption{Probability tree for the CP-logic rule, illustrating the possible outcomes and their associated probabilities (image inspired by~\cite{vennekens2009cp}).}
\label{fig:cp-logic-tree}
\end{figure}
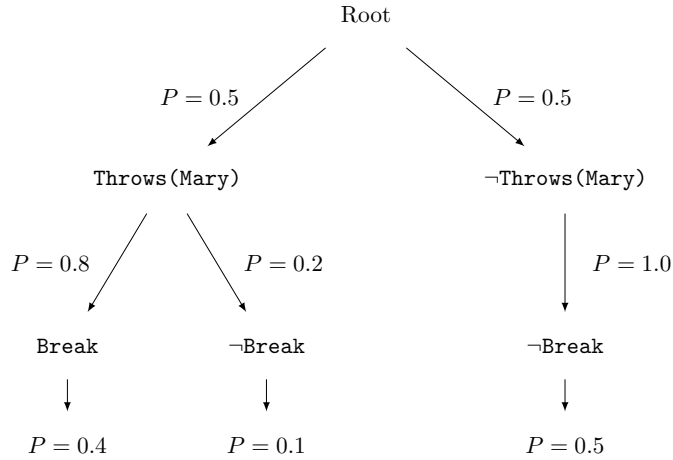

For example, consider a simple (ground) probabilistic causal law:
\[
\texttt{(Break : 0.8)} \leftarrow \texttt{Throws(Mary)},
\]
indicating that if Mary throws a rock, the window breaks with probability \(0.8\).
The corresponding tree is then shown in Figure~\ref{fig:cp-logic-tree}.

The execution model of a CP-theory is then a probabilistic process consistent with the specified causal laws. For a given set of probabilistic causal laws, CP-logic ensures that all valid execution models yield the same unique probability distribution over final states. This is an important property for predictive consistency, demonstrating the usefulness of causality in describing the class of probability distributions that can result from a given probabilistic process, while allowing to ignore irrelevant details (e.g., the name of the person throwing a rock).

Formally, let \(C\) be a CP-theory, and \(X\) an interpretation of exogenous predicates. A probabilistic process \(T = \langle T; I; P \rangle\) is an {execution model} of \(C\) in context \(X\) if:
\begin{enumerate}
    \item \(I(s) = X\) at the root \(s\) of the tree.
    \item For each non-leaf node \(s\), a CP-law \(r\) fires, such that its precondition \(\phi\) is satisfied: \(I(s) \models \phi\).
    \item For each leaf node \(l\), no CP-law has its precondition satisfied: \(I(l) \not\models \phi \text{ for all } \phi.\)
\end{enumerate}
Under these constraints, CP-logic provides a structured way to model causal relationships in dynamic systems, offering a compactly lifted~\cite{kimmig2015lifted} representation. 
Compared to the Bayesian networks, the use of logic in CP offers a more modular, qualitative, and general framework for representing probabilistic causality (e.g. in representing cyclic causal relations and more complex mediation structure to variables' influence on each other).

\paragraph{Other Causal AI Formalisms}
Beyond SCMs and CP-logic, various alternative approaches have been proposed to address causal reasoning in AI, particularly in the domain of ``actual causality''~\cite{halpern2016actual}, also referred to as \emph{token causality}. In contrast to the aforementioned views, building mostly on Pearl's ideas, the concept of token causality, developed mostly upon the work of Harpen~\cite{halpern2016actual}, seeks to identify specific causes for particular outcomes in individual instances. As opposed to the generalized type-level causal relationships used in machine learning, the token causality is more concerned with formal reasoning and the goal to axiomatize the intuitive form of causal reasoning as used by humans~\cite{kueffner2021comprehensive}.

Token causality is most prominently formalized through \textit{counterfactual reasoning}, as originally advanced by Halpern and Pearl~\cite{halpern2005causes}. These theories define causation in terms of hypothetical ``what-if'' scenarios. Specifically, a variable \( A \) is said to be a cause of \( B \) if, in a model \( M \), setting \( A \) to a different value changes the outcome \( B \). Formally, let \( M \) be a structural causal model and \( X \) a set of endogenous variables with structural equations \( F \). For \( A = a \) and \( B = b \), \( A \) is a cause of \( B \) if:
\[
M, A = a' \not\models B = b \quad \text{for some } a' \neq a.
\]

Following a recent comprehensive survey~\cite{kueffner2021comprehensive}, one may categorize the actual causality approaches based on their philosophical foundations~\cite{beebee2009oxford} as follows.

\textit{Regularity theories} link causality to patterns of occurrence, often aligning with type-level causal models. 
\textit{Probabilistic theories} further extend this perspective by asserting that \( A \) causes \( B \) if the probability of \( B \) increases given \( A \):
\[
P(B \mid A) > P(B \mid \neg A).
\]
\textit{Causal process theories} then focus on continuous mechanisms that connect cause and effect, emphasizing the physical transmission of influence. These models are less suited for discrete event systems but are applicable in domains such as physics and biology where dynamic interactions dominate.
\textit{Interventionist theories} emphasize causality as the effect of deliberate manipulations, formalized through Pearl’s do-calculus. 
\textit{Agency-based extensions} frame causation in terms of an agent’s ability to influence outcomes, making these approaches particularly relevant for reinforcement learning and decision-making applications.

Additionally, one may also evaluate token causality models based on their applicability across different benchmarks, including classic challenges like preemption, overdetermination, and switching scenarios~\cite{kueffner2021comprehensive}. For instance, the CP-logic~\cite{vennekens2009cp} is particularly adept at handling preemptive causation, whereas counterfactual models excel in straightforward dependency structures.

Defining reasonable notions of actual causality is challenging to comprehensively axiomatize. The fact that this program has not converged, and diverse scenarios continue to present either the inadequacy or insufficiency of any singular set of such axioms, suggests this method of performing the Philosophy of causation will not converge. We can observe, however, the important role that intuition and thought experiments have played in the rigorous testing of correspondence between causal axiom sets and ``real world causal claims''. This natural tendency to get to the bottom of abstract concepts provides impetus for the use of Ordinary Language Philosophy, i.e. elucidation through precise Semiotic anthropological description of how the concepts are used in practice. 

\subsection{Background: Neuro-Symbolic and Explainable AI}

AI paradigms can be broadly categorized into ``symbolic'' and ``subsymbolic'', as rooted in the historical developments in the AI discipline. Generally, symbolic AI methods rely on qualitative formal knowledge representations, while the subsymbolic, often referred to as ``statistical'', AI methods focus on quantitative data-driven modeling.
The currently dominant neural networks naturally fall into the subsymbolic (statistical) category, due to the numeric distributed nature of their internal representations (embeddings), and their purely data-driven training procedures.


\paragraph{NeuroSymbolic AI}
Recognizing the complementary strengths of the symbolic and {neural} approaches, many researchers have pursued ``neural-symbolic integration'', aiming to combine advantages of both~\cite{besold2021neural} - the controllability and interpretability of symbolic systems with the robustness and scalability of neural networks. Following a taxonomy similar to Kautz~\cite{kautz2022third}, neuro-symbolic systems can be roughly categorized based on their level of integration. 
The first category involves independent assemblies of symbolic and neural modules, such as a deep learning system utilizing a symbolic knowledge base for input preprocessing or a symbolic search algorithm guided by a deep learning heuristic estimator. 
The second category features a tighter coupling, where the neural network principles integrate directly with some form of symbolic knowledge representation, such as first-order logic~\cite{sourek2018lifted}. Many systems fall in a grey area between these categories, exhibiting partial levels of integration. 
Finally the third category, closely related to XAI, encompasses symbolic surrogates for neural models and methods for extracting symbolic information from trained networks to facilitate \textit{explainability}.

\paragraph{Explainable AI}

Explainable AI (XAI) describes methods that aim to render the workings of AI systems more transparent and understandable to humans. This is essential not only for building trust in the models' decisions, but also for regular debugging, improving, and gaining insights from these systems. XAI encompasses a variety of techniques, ranging from statistical feature importance analysis to more involved methods like LIME (Local Interpretable Model-agnostic Explanations)~\cite{ribeiro2016should} and SHAP (SHapley Additive exPlanations)~\cite{lundberg2017unified}. These methods seek to provide explanations for \textit{specific} predictions made by a model, either by approximating the model locally with a simpler, interpretable one (LIME) or by assigning importance values to features based on game-theoretic principles (SHAP). XAI also sometimes addresses global interpretability, aiming to understand the overall behavior of a model, rather than individual predictions. Methods like decision trees and rule lists offer inherent global interpretability out of the box, but their application is often limited by the complexity of the domain. One approach associated to one of the formalisms for causal inference, counterfactuals, is popular in XAI -- the user is assisted in understanding the classification label by offering counterfactual explanations indicating which changes in the input would have led to a different prediction~\cite{wachter2017counterfactual}. 
The effectiveness of XAI depends on fidelity, comprehensibility, and relevance, with the choice of method guided by application, model characteristics, and user needs~\cite{adadi2018peeking}. Importantly, this choice introduces an inherent inductive bias, shaping the explanation and thus the eventual (causal) user understanding.

\paragraph{Neuro-symbolic XAI}

Neuro-symbolic XAI seeks to combine existing strengths of neural networks (e.g., robustness, scalability) with symbolic reasoning (e.g., interpretability/explainability) to develop XAI methods with richer explanatory power w.r.t. the underlying (causal) mechanisms. One of the original, and still prominent, approaches is to extract symbolic representations (e.g., logic rules, semantic graphs) from trained neural networks~\cite{hailesilassie2016rule}. This can provide higher-level insights into the model's learned behavior, going beyond feature importance, to capture relationships and underlying reasoning processes~\cite{besold2021neural}.  Another approach is to integrate symbolic knowledge directly into the learning process, guiding the neural network to learn representations consistent with some symbolic prior knowledge~\cite{sourek2018lifted}. This can improve both the generalization and interpretability of the resulting model. Neuro-symbolic XAI has a potential to address the limitations of the purely statistical XAI methods, particularly in complex domains requiring a deeper understanding of the underlying causal mechanisms, but challenges remain in integrating these disparate paradigms effectively for managing the classic complexity-explainability trade-off.


\paragraph{Mechanistic Interpretability}


Mechanistic interpretability is an emerging field that, in a spirit reminiscent of the neuro-symbolic XAI, aims to understand the internal mechanisms of neural networks, with a particular focus on Transformers and language models. It is a set of techniques in which AI researchers attempt to model and apply scientific practice to understand phenomena in the natural world.

A neural network is most commonly a function \( f: \mathcal{X} \to \mathcal{Y} \), transforming an input \( \mathbf{x} \in \mathcal{X} \) into an output \( \hat{\mathbf{y}} \in \mathcal{Y} \) through a series of intermediate transformations:
\[
\mathbf{h}_i = \sigma_i(\mathbf{W}_i \mathbf{h}_{i-1} + \mathbf{b}_i), \quad \hat{\mathbf{y}} = \sigma_n(\mathbf{W}_n \mathbf{h}_{n-1} + \mathbf{b}_n),
\]
where \( \mathbf{W}_i \) and \( \mathbf{b}_i \) are weights and biases, and \( \sigma_i \) denotes elementary activation functions. Mechanistic interpretability generally seeks to decompose \( f(\mathbf{x}) \) into sub-functions \( f_i(\mathbf{x}) \), representing functional units or \textit{circuits}~\cite{olah2020zoom} responsible for specific behaviors:
\[
f(\mathbf{x}) = \sum_i f_i(\mathbf{x}).
\]

Key techniques include tracing information flow through layers and analyzing circuits~\cite{olah2020zoom}, leveraging tools from network science, information theory, and causal inference. For example, network science methods identify important neurons or sub-networks through connectivity analysis, while information theory quantifies mutual information \( I(\mathbf{x}; \mathbf{h}_i) \) to assess how specific neurons retain relevant information~\cite{shwartz2017opening}. Causal inference techniques, such as counterfactual reasoning, can then be used to explore how perturbations to different parameters \( \mathbf{W}_i \) influence outputs:
\[
\hat{\mathbf{y}}_{\text{cf}} = f(\mathbf{x}, \mathbf{W}_i \to \mathbf{W}_i').
\]

Most commonly, sparse autoencoders~\cite{cunningham2023sparse} are then used to further aid interpretability by enforcing sparsity in the learned representations through solving:
\[
\min_{\mathbf{W}, \mathbf{b}} \| \mathbf{x} - \sigma(\mathbf{W} \mathbf{x} + \mathbf{b}) \|_2^2 + \lambda \|\mathbf{W}\|_1,
\]
to highlight the most critical features. This technique encapsulates a general theme of algorithms and models purported to identify cause and effect as considering the discovery of sparsity in the model. We shall see this set of tools can be understand as statistical method that attempts to extract model information that can be useful for scientific causal investigation, even if it would be incorrect to say they definitively identify and certify these investigations.

\section{Causality and Epistemology}
Causality is a central theme of philosophy, of importance for meta-inquiries of existence and knowledge. For instance, the Metaphysics of the ancient Greeks is often described as a general fundamental ontology causing circumstances and observations of the world around us. The distinction between the Natural Sciences and Metaphysics was hardly present in early philosophical methods, and the general exercise of Natural Philosophy purported to define the fundamental forces that govern the processes of the world.

The question of how one can know what caused a particular event that is, inquiries of the form ``how does one know for certain that A caused B?'' are a theme throughout philosophical inquiry. Here, causality itself is typically taken as an intuitive notion that must have a central validity in any sort of sensible model of the world. That is, rather than starting with a formal definition of ``cause'', the concept is instead used in philosophical treatises as a commonly understood notion. It is then used, with certain implicit central characteristics of the concept of causality, as a tool for argument or consideration, that is, adjudicating philosophical meta-narratives by testing their application to causal semantics. 

We briefly summarize the primary historical developments on the ideas surrounding the question of how to ascribe knowledge and certainty to claims of cause of effect, and the different answers to it. Noting harmony with themes of the omnipresence of inductive bias in Contemporary Philosophy, particularly in Quine and Kuhn, we continue to focus specifically on the semantics of ``cause and effect'' through the Ordinary Language Perspective, as espoused in \emph{The Philosophical Investigations}~\cite{wittgenstein2009philosophical} and other works. We conclude on considering the role of Hermeneutics, as the philosophy of understanding and thus the appropriate method for conceptualizing interpretability. Ultimately, any statistical model, if claiming a cause and effect process, should be understood as such by someone in the associated domain. The multiplicity of scientific domains and methods of practice indicates the necessity of considering causality itself as a generic, or lifted, concept, rather than one with its own essential semantics.

\subsection{Early Work}
The first antecedents of causality in the sciences lie in various cultures' treatment of mind, emotions, and the soul. They variously classified the components of subjective experience and their relationships, and attributed these components and their relationships to different parts of the body (as in Hippocrates' theory of humors and Galen's medical writings). Descriptions of internal and external processes had a greater emphasis in early cultures (relative to ours) on a prominent and complex external origin, frequently divine, for thoughts, inspirations and insights, with divine communication of the nature of reality coming particularly through dreams and visions. The foundational mechanisms and central processes of the world were frequently thought to correspond to a separate, more perfect world of true shapes and functions (as in Plato's Theory of Forms). HoweverOutside of western traditions, Indian philosophical schools, particularly Buddhist theories of causation like prati tyasamutpada (dependent origination) and the Nyaya school's sophisticated theories of perception and inference (pramaṇa), developed complex frameworks for understanding mental processes and consciousness. Similarly, Chinese philosophical traditions, particularly in Neo-Confucian thought, developed sophisticated theories of mind-body relations and causal reasoning.

The transition to early modern philosophy was bridged by crucial medieval developments, particularly from Islamic philosophers. Avicenna's (Ibn Sina's) floating man thought experiment and theory of intellect, along with Al-Ghazali's skeptical approach to causation in ``The Incoherence of the Philosophers,'' significantly influenced later Western thought on mind-body relations and causal necessity. Around the time of Descartes, Hume, Berkeley, et cetera (17th-18th centuries), a series of philosophical thought experiments shed doubt on the notion that there was an objective world that was being merely apprehended by the organs of the senses. Furthermore, the substrate of the mind came to be less diffusely attributed to organs or humors, and came to be considered increasingly either a function of the brain or the function of a substrate entirely separate from matter (Cartesian dualism). 

\subsection{Causality, Empiricism and Dualism}

One argument of Descartes for the existence of God went as, basically: every event must have a ``mover", we can consider a chain of events going increasingly back in time, and so if we continue this process who was the first mover? One could readily identify mover as generically meaning a physical cause. That is every event has a cause with implicit temporal backtracking, and broadly, chaining of the sequence of events. So it is visible that the notion presupposes certain axiomatic definitions with respect to the presence of causal phenomena and realities. Furthermore, Descartes insisted that that everything \emph{must} have a cause, as a claim of a priori fact as to the fundamentals of the universe. This is a classic example of \emph{a priori} reason, that is, claims of knowledge that are not tautological derivations of existing facts and not based on empirical observations.

At the same time, one can note the fact that the arguments put forth by Descartes were broadly respected in the intellectual landscape of the time, and suggest that the notion of a``mover'', according to the way Descartes uses the term, is intuitively equivalent to what people commonly think of as a cause. This indicates salient features of the concept of cause and effect: 1) that it is an abstract and generic concept that can be universally understood through knowledge of language alone, rather than requiring education and expertise, 2) that it is a central component in the model with which humans understand and process the world, and 3) that certain axioms regarding cause, that of being omnipresent in every phenomenon and following temporal sequence, are intuitively self evident. These observationsprovide clues as to the possible correspondence between certain patterns of formal abstraction in scientific practice and the use of cause and effect in common everyday speech.

With Descartes setting the program of western early modern Epistemology, Philosophy continued to develop towards developing a research program of attempting to refine and precisely formalize belief and knowledge, with causation as a regular theme.
David Hume was particularly frustrated with the failure of induction to provide certainty in evidence for claims, broadly speaking, with confident knowledge in ascribing causation being a central element of his writing. In particular, he emphasized the limitation in reason that it can happen that one may observe event A happen before event B a great many times, and yet such data by itself cannot ensure that A causes B with complete certainty. For instance, it can happen many times that lightning precedes thunder but no matter how many instances this is observed, one cannot conclude that lightning causes thunder. Here, cause is again still taken as a self-evident notion that presents a central role in connecting two events. His attribution of truth as either synthetic and empirical, that is new facts drawn from observations of the world, or analytic and rational as tautological derivations, leaves causation in a fundamental haze of uncertainty. 

Immanuel Kant's Epistemology is typically considered the climax of early modern Western philosophy, with strong claims of precise delineation of forms of knowledge. \emph{Critique of Pure Reason} wrestles, through inquiry, thought experiments, and models of truth and knowledge, with both the practical necessity for, and the fundamental challenges in confirming, claims of fact arising purely from \emph{a priori} theorizing. David Hume's skepticism in regards to cause and effect is answered with the assertion that the notion of causality is \emph{a priori} knowledge that comes from intuition in the mind that is fundamentally true. The sense world, that is, what is observed, is then fit into this general framework of world understanding. This presents causality as a lifted system of knowledge with its own logic, fundamentally embedded in human cognition. 

\subsection{Falsifiability and Paradigms}

Karl Popper's grand thesis in his seminal work on the philosophy of science relies on the notion of falsifiability ~\cite{popper2005logic}. Here a statement is a valid scientific theory or law only if it can be translated into a practical experiment, measured and possibly rejected if the observed data are not consistent with theoretical predictions. In line with American Pragmatist traditions~\cite{peirce1972charles}, Popper rejected the possibility of \emph{a priori} statements of fact, and as a solution to Hume's concern that induction can never lead to knowledge of certainty, asserted that any scientific claim should be treated as potentially fallible owing to the possibility of future empirical rejection. Here, to ask if one phenomenon causes another requires a scientific theory that credibly describes this causation and implies possible experiments that could test the claim's validity. An extensive set of empirical data that is consistent with the theory supports elevating a hypothesis or conjecture into a law. Meanwhile, data that contradicts an implication of the causal claim should be treated as a rejection of the idea altogether, noting that good faith science requires the ability to formulate such potential experiments.

Many disciplines, however, fall short of this requirement. In sociology, for instance, controlled experiments are impossible. In psychology, even the most prominent and prolific theorists would consider it delusional to claim to have a theory that can predict an individual person's behavioral response to some external condition with absolute certainty. Rather, the most that is claimed are tendencies that can be observed in broad statistical patterns. Theories in these fields typically do not make definite precise quantitative predictions, nor can they. Psychometrics and Econometrics are well developed statistical toolkits for analyzing data from social domains, but the significance tests they undertake do not have a clean one to one correspondence to definite clauses of Psychological or Economic fact. Yet, despite these realities there are still well respected and reliable fundamental theories of cause and effect in these fields. 

These violations of standard requirements for asserting cause would seem to invalidate the entire practice of fields including psychology and sociology, as well as anthropology, political science, and even economics. Yet, psychology has yielded therapy that is beneficial to human flourishing, and economics can yield macroeconomic policy that brings greater wealth, prosperity, and stability thereof. It appears that while fallibility and falsifiability are theoretically consistent as a Philosophy of Science, they are insufficiently useful.

A more powerful and fundamental critique comes from Thomas Kuhn. In the seminal work \emph{The Structure of Scientific Revolutions} he posited that falsifiability itself depends on a certain background paradigm to indicate when a given observed phenomenon is even potentially fallible. That is, the experimental device and measurement system depend on systems of assumptions regarding the mechanisms and dynamics of the world, i.e. paradigms. These paradigms are ultimately axiomatic, unfalsifiable, and even prelogical, and present a global scaffolding on which scientific practice develops. Occasionally, an anomaly appears, which is typically dealt with by some ad-hoc modification of the theory. As anomalies accumulate, an opportunity arises for a scientific revolution in which a new paradigm supplants the old one, simultaneously fiting the previously observed phenomena as well as the contemporary anomalies.

As an example of this social process in even the ``hardest'' of science, physics, see~\cite{pickering1986constructing}. Here, one can readily see how subjective choices based on biases towards elegance and sunk-cost-fallacy inertia influenced research on quarks as the fundamental building block of particles and their triumph over competing paradigms. This subjectivity in practice does not invalidate the enterprise, as some antiscientific radicals may assert. Rather, quarks are the governing paradigm, and the gauge conditions and structure of the field equations they imply is consistent with experimental results in particle physics as well as other scientific domains concerned with more curtailed sets of phenomena, such as electromagnetism. Science proceeds as according to the quark ontology, and it should, as it is the most complete and precise model available, and, to the best of the authors' knowledge (who are not, however, particle physicists, so this statement is not authoritative), no sufficiently concerning accumulation of anomalous experimental observations have appeared. 

Imre Lakatos' work (e.g.~\cite{lakatos2019and}) focuses on reconciling Popper and Kuhn. He posited that paradigms can be described  as a logical hard core, a central set of theoretical systems which defines the basic principles and is itself subject to intellectual development via peer scrutiny. Then, auxiliary hypotheses stand on top of these, and can be progressive or degenerate depending on whether they are productively developing a superstructure with a potential to generate new predictions or using ad-hoc solutions to solve anomalies. Ultimately, we shall see that ascribing cause and effect depends on the central mechanisms of influence defined in the hard core of a scientific paradigm. Here, theoretical and empirical investigation as to what causes what effect in the domain of interest, are formed within the structure of the representation defined by domain's language of causality. Thus, one cannot divorce questions and answers regarding causal phenomena from the the hard core of a scientific paradigm. 

Before teasing out the implications of this inescapable complexity for causal learning, we must review Quine's work in Epistemology.


\paragraph{Quine and Inductive Bias}
W.V. Quine's classic essay on epistemology, the \emph{Two Dogmas of Empiricism}~\cite{van1976two} is a seminal contribution on role of inductive bias as underlying any understanding of causality. The two dogmas that Quine resolutely discredits are the sharp delineations between synthetic and analytic knowledge, i.e. immediate and self-evident new knowledge compared to that derived from a tautological process of observation and/or reasoning; between empirical and rational; or the classic dualism of performing inferences from sense data observations vs. through abstract reasoning and deduction. The long-held standard of a theory's potential falsifiability through empirical observations as the only reliable means to adjudicate scientific claims ultimately and intimately rests upon the epistemological foundation of these two dogams.

He argues and demonstrates that a far more accurate and sensible map of knowledge is as that of a web. With a resemblance to the notions of a hard core and auxiliary periphery of Lakatos, however with a more graduated set of connections of derivation and subsumption rather than crisp demarcations, the web has a dense highly connected center and looser exterior. Observations and experiments can be defined as a new connection of a prediction of the theory in the specific experimental setting. The web is extended at the periphery as based on the real world predictions the model provides. The outcome of the experiment can then potentially falsify this prediction. That is, this is evidence \emph{given} assumptions of more fundamental theoretical constructions, that are recognized as contemporaneously valid. Since the prediction only makes sense within the context of the web around it, the choice of which connection of the web the experiment invalidates is underdetermined. As a matter of efficient scientific practice, the nodes that are most peripheral and with the fewest total connections should be the first targets of potential incision. These will, of course, constitute the more peripheral scientific knowledge rather than the hard core, and so treating anomalies as simply an opportunity to refine and patch exceptions to a reigning paradigm, rather than as necessitating scientific revolution. That revolutions are risky and involve an immediate drop in scientific output per worker are concerns that can be mitigated with proper planning and organization. 

The Logical Positivist dualist interpretation of experiment, as synthetic a posteriori data, and logical deduction and mathematical derivation from one theory to another as synthetic a priori knowledge, corresponds to a simple crisp structure for adjudicating causal claims. Consider again causal learning as defined by the textbook Markovian probability kernel defining a (Dynamic) Bayesian Network. By this framework, a set of statistically significant conditional independence tests can confirm a model of the causes and effects of variables. However this, by itself, is not well defined. One has to formulate a specific transition kernel, such as a Generalized Linear Model or Linear Structural Equation Model. One can then inquire as to the basis for one choice of such representation or another, and ultimately finds that the choice is inductive bias. In addition, there are more fundamental questions as to the validity of Bayesian networks in the first place: Why are the dynamics purely Markovian? Why can there not be cyclical patterns of causation?  




In the grounded conduct of scientific research, an experimental result is, in practice, generally used to either 1) calibrate and quantify an explicit mathematical model or 2) perform some hypothesis (so-called ``AB'') test. For both tasks, the empirical inquiry is intimately tied with some theoretically grounded inductive bias. Calibration involves, typically, solving inverse least squares problems wherein a parameter-dependent system of equations is forward simulated and observational data is used to define an optimization problem whose minimization corresponds to a maximum likelihood estimate for the data given the parameters.   
In the second case, the likelihood of two models is compared to consider the relative credibility of each. This choice of formalism for the forward simulation, defined as the \emph{model representation}, is the inductive bias present in these lines of research. Beyond this, the choice of relevant and irrelevant variables to measure and include into the statistical model in the first place is driven by the associated inductive bias. 

In all but simulated and highly controlled industrial physics conditions, any phenomenon will have some interaction with a range of objects and forces in the environment that would be even fundamentally impossible to formally account for, thus practically necessitating choosing the most relevant variables, which is a methodological choice of practice rather than one with a clear best answer. Returning to the graphical conditional independence paradigm, we have that there is a condition of \emph{causal sufficiency}, that is that all causal sources are present in the measured variables, that is assumed formally. Although this appears to be standard methodologically, in practice this assumption is unrealistic. \footnote{This note finds correspondence in some observations appearing in causal machine learning: sparse disentangled models have been observed to only be identified in the presence of inductive bias~\cite{locatello2019challenging}.}

\subsection{Ordinary Language}
Ordinary language as introduced by Wittgenstein~\cite{wittgenstein2009philosophical} presents a standard methodology of performing Semiotics in regards to philosophical concepts, which we can employ by studying the language games under which ``causality'' and ``cause'' are used. They are used in both common parlance as well as scientific language, which presents an opportunity for confusion. To begin with, let us consider scientific context, in order to more easily align any potential correspondence or lack thereof with the Classical Causality Paradigm.

\emph{Scientific Use of ``Causality''}
Considering the vastness of the setting, from ``heat causes ammonia to react'' to
``overstretching caused the fall of the Roman Empire'', it is clear that the use of the word is conceptual, and broadly applicable to any environment and setting wherein things interact by mechanisms that could be described by experts in a manner that is mutually understood and agreed within the paradigm. One could clearly indicate, for instance in the two examples, that a chemist can explain how heat causes ammonia to react and a historian of antiquity would be able to explain a sensible narrative of the Fall of the Roman Empire. 

Every scientific discipline, practice and set of institutions has specific language games which characterize appropriate and successful operation of dialogue in that scientific context. 
When one speaks of causes, then, it is clear that one speaks of a fundamental grammatical category in a particular scientific language game. 

From an anthropological perspective, each discipline and its associated language game(s) can be defined as a tribe,  with its particular customs and rules of conducting scientific inquiry. We can thus see that ``cause'' is in a sense universal, that is, there is something that corresponds to a notion of cause across disciplines, though the specific language game used is distinct in form. 

This is clear with a casual Anthropological impression of scientific discourse. Physics is characterized by field equations, and an evolution equation with a potential term is a canonical mathematical structure defining cause and effect. In addition, experiments can be performed under very specific experimental conditions that are effectively closed to external environmental influence. A 5$\sigma$ significance is standard in particle physics. 

In contrast, in Medicine, causality is often associated with a Randomized Clinical Trial. The quintessential case wthin observational studies is with the health effects caused by smoking.In this case, properly independent statistical significance many times over a variety of data sources was noted together with an entire mechanistic narrative tar build up leading to arterial clogging and lung damage. This telling example presents a methodological agglomeration of a combination of evidence types towards a narrative of cause-effect relationship between variables. 

In the social science, ``causes'' are often complex narratives. Grand claims of causation are often entire seminal treatises. We can take \emph{The Protestant ethic and the spirit of capitalism} as an example. The claim, that Protestantism endowed people with certain character traits that made the adoption and ability to excel at Capitalism, defines a path by which religious practice caused aggregate group behavior which caused prosperity. 

Social sciences can provide insight and meaning into human life and activity, despite the relative lack of rigor and counterfactual nature compared to the physical sciences. Indeed, narratives are a standard mechanism by which humans process meaning in the world. Thus, while one cannot make exact quantitative inference, even with margins of error, of sociological and psychological phenomena, one can still present an understanding of the possibilities and the way different phenomena can be comprehended within the context of one's life. For example, there is an accumulating evidence that while, statistically on average, psychotherapy is effective in improving mental health conditions, the therapy modality does not affect the success rate~\cite{echeburua1996comparative}. Given that the modality used (CBT, psychodynamic, etc.) is founded on fundamentally different conceptions of the human mind with different causal narratives, it is clear that rather than empirical tests, the priority is for the narrative of the therapeutic theory to be compelling to the practitioner. 

Finally, let us remark on the use of ``cause'' in common parlance. One can note that this is typically either a lay simplification of a scientific statement, such as a particular billiard ball knocking another one into a hole being a statement of classical mechanics, or political statements regarding macroeconomics. Other discussion can be about tactics or strategy, e.g., regarding sports. These descriptions are sufficiently diverse so as to frustrate any attempt to formally axiomatize them (despite attempts~\cite{vennekens2009cp}). In addition, as the language game involves narratives that can be rather flexibly applied, this presents the necessity of vigilance. For instance, it can correspond to the purely social legal concept of fault, ``he caused this crime to occur'' is an indication of a juridicial notion of guilt. We can see the potential for ideological influence in this context, as well as both misguided and even tactical~\cite{orwell2021politics} uses of language of obfuscation to, e.g., accuse certain people and institutions of certain activities or even intent.


\subsection{Hermeneutics and Causal Narratives}

Considering the contingency of causal descriptions on the specific grammar of scientific language games presents natural Anthropological questions. Findings need to be communicated in a proper way to those interested, including other researchers, policymakers, and laymen. Here \emph{Hermeneutics}, as the field of understanding how humans process meaning, presents itself as the proper framework for studying these considerations. Hermeneutics can be understood as being strictly in between the layers of abstraction that are purely hard scientific (e.g. articles in physics journals) and those that are purely aesthetic (e.g., literature). This occupying of a liminal place in the hierarchy of verity can be motivated by the observation that stories are compelling to humans, and through their visceral appeal, can induce changes in behavior, and of group collective behavior, e.g., voting. Thus, there is still an \emph{instrumental} meaning to narratives as far as they, in some sense, cause behaviors, that is not as exact as statements of physical processes but not as purely recreational and sociocultural as poetry. 

In \emph{Truth and Method}~\cite{gadamer2013truth} Gadamer outlines the principles of Hermeneutics and contrasts it to contemporary social science and philosophy. He argues two fundamental features of understanding and interpretation that are particularly relevant for our exposition in this paper: 1) the epistemologically flawed approach of social science trends toward seeking to resemble practice in the natural science, and, 2) that there are structural universal characteristics and properties of how people understand, that is, Hermeneutic truth can be studied in the abstract, yielding lifted insights that apply cross-language games, i.e., beyond contextual representations.


With this, we can see that causality-as-interpretation becomes a model that is applicable across the scientific domains, and thus is a parsimonious explanation of the universal use of words like ``cause'' and ``causation'' across scientific fields, despite their fundamentally different form and semantics when it comes to the language of theory in each domain. This explains how a differential evolution equation in physics can be understood as defining cause and effect in the same way as a treatise on the details and history of some specific recent cultural phenomenon. We shall explore the details of these disparate methodological practices below as we survey some of the disciplines. 

Note that fundamentally the graphical conditional independence model of causality becomes problematic not insofar as it is formally invalid, although there are clear instances that one can claim as such even within the framework of formal causal learning axioms (i.e., causal sufficiency is never really satisfied in practice). Rather, it is an inappropriate choice of level of abstraction, and amounts to redefining the word ``cause'' to be some specific statistical machine learning tool. Yet the word ``cause'' still has a real meaning within any scientific discipline, one that is critically important to its canon of work. This can present considerable confusion to the point of introducing errors in claims or the confidence therein with a methodological incorporation of the machine learning technique specifically into any scientific discipline, the exact ``bewitchment of intelligence by language'' that clear philosophical investigation is meant to fix. 

\section{Semantics Across the Sciences}
\subsection{Physics}
In Physics there are mathematical foundations that completely and comprehensively express theoretical physics' model of the world. That is, while narrative explanations of physical intuition are helpful for understanding, all assertions regarding cause and effect in the physics domain can be expressed in a formalism of equations without any essential component missing. There is a finite set of potential forces that can be exerted, four considered fundamental and the rest derivative.  Certain canonical field equations represent the balance of forces on various given types of systems and environments. For instance Einstein's Field Equation for General Relativity,
\[
R_{ab}-\frac{1}{2}R g_{ab}+\Lambda g_{ab}=\kappa T_{ab}
\]
where $R_{ab}$ is the Ricci Curvature, $R$ is the Scalar Curvature, $g_{ab}$ is the space-time metric, and $T_{ab}$ is the stress-energy tensor of mass and forces present in spacetime. The ontology of General Relativity is that $T_{ab}$ influences the very geometry of spacetime in such a way as to make objects, moving along straight paths on curved space, end up moving towards the direction of force, defining the mechanism of gravity. That is, the field equation are completely sufficient to describe the force of gravity, and any statement in the form of, for example, ``the gravitational pull of Planet Z caused the moon to orbit elliptically'' can be expressed, potentially, as specific expressions for the stress energy tensor defining Planet Z's mass density and the metric $g_{ab}$ or derivative tensor expressions of its quintessential geometry (e.g. Christoffel symbols). Although physical explanation assist understanding, there is a one to one correspondence between a fully specified causal physical scenario and a set of equation systems. 

Differential equations, either ordinary or evolution (time-dependent) partial differential  equations describe the change, often acceleration, of something as a function of certain modeled quantities. For instance the presence of electromagnetic potential accelerates a charged particle, elastic stress deforms the shape of a material continuum, etc. In the case of General Relativity, it is common to perform the 3+1 foliation and consider a set of time slice constraints and an evolution system defining the change in the spacetime metric with time. One can infer consistent patterns and relationships between quantities through simulating the stationary solutions, the so called Einstein constraints, and define the mechanism of change through the force term in the evolution equations. This semantics of causality in physics gives it the unique precision in the sciences that it enjoys. There is no effect or activity that does not have an essential corresponding expression that fits with the hard core of that particular physical domain (General Relativity, Electromagnetism, Statistical Mechanics, etc.)

More generally, ordinary and evolution Differential equations translate to ``the change in one quantity with respect to time as a function of the value of some other quantities,'' which can be readily interpreted as the quantities in the function causing changes in the former. In a set of field equations, one can consider that there can be multiple systems all interacting with each other, such as two magnets near each other, and thus a change in the state of any one would be understood to cause a change in the other. For instance, consider solving the homotopy, from $\tau=0$ to $\tau=1$, of Maxwell's equations with an external charge applied with strength $t$. Maxwell's equations, which solve for the magnetic field $\mathbf{B}$, the Electric Field $\mathbf{E}$ given a current $\mathbf{J}$ and, for this example, a potential function $V$ defining the interaction of two magnets $(\mathbf{q}_1,\mathbf{q}_2)$.
\[
\begin{array}{l}
\nabla\cdot\mathbf{E}=\frac{\rho}{\epsilon}\\
\nabla\cdot\mathbf{B}=V(\mathbf{q}_1,\mathbf{q}_2)\\
\nabla\times\mathbf{E}=-\frac{\partial\mathbf{B}}{\partial t} \\
\nabla\times\mathbf{B}=\mu\left(\tau\mathbf{J}+\epsilon\frac{\partial\mathbf{E}}{\partial t}\right)
\end{array}
\]
with parameters: 1) $\rho$ the electric charge density 2) $\epsilon$ the permittivity coefficient, 3) $\mu$ the permeability coefficient.

As $\tau$ increases, the current $\mathbf{J}$ increasingly exerts an influence on both the magnetic and electric field in the fourth equation, the Amp{\` e}re-Maxwell Law. This law, which establishes a significant coupling between the electric and magnetic field, and is in fact responsible for electromagnetic waves, can result in profound changes in the overall system as caused by the injection of the current.

Thus, causation can be understood as sensitivity perturbation results on the equations corresponding to the system, that is, e.g. $\frac{\partial \mathbf{B}(\tau)}{\partial \tau}$, the change in the solution with respect to the parameter. An intervention can also be considered as an external force that makes the system lose equilibrium and transition to a different state. This transition can be said to have been caused by the intervention external force.

As far as techniques associated with machine learning, it has been argued that physics-informed neural networks already infer causality with certain dynamic equations enforced in the system~\cite{wang2022respecting}. Indeed, the practice of using the Hamiltonian or Lagrangian as a conserved quantity of physics-informed neural networks is in line with the understanding of causality, or the physical mechanisms of any phenomenon, as modeled by field equations. Note that this clearly highlights that \emph{only when} the empirical machine learning tool directly fits the representation that is present in a scientific domain, does empirical and statistical inquiry directly assess causal claims with respect to some formalized equation system.

\paragraph{Engineering As Applied Emergent Physics}
In Engineering one can say that a change in one component of a system could have caused some change in another. For instance, the breaking down of a vehicle caused by a drivetrain problem. Better acceleration caused by deeper cylinders. More sophisticated gyroscopes causing a robot to navigate better. As Engineering could be considered applied physics, one can say that in this case the physics of the system is understood, and there is a correspondence between the mechanisms within a system to different states of emergent behavior of the system as a whole. Certain unstable phenomena inside a component can result in the entire system malfunctioning, necessitating the vigilance that precise physical equation modeling can provide. On the other side of this aspect is that a synergistic interaction between two components can cause it to ``perform'', in some appropriately measured sense (e.g., produce a certain amount of energy), better.

As such, engineering can be described as a \emph{clinical} branch of Physics. We can observe that this label harmonizes with the core-periphery model of knowledge as developed in 20th century Epistemology -- Physics is a paradigm with a set of more or less uncontested sets of principles and laws. Engineering narrows the context towards more specific processes, permitting acceptance of greater flexibility in adjusting models to empirical evidence. Moreover, \emph{how} models adjust to evidence is itself a well respected branch of, so-called \emph{computational}, physics, including the deep fields of \emph{Uncertainty Quantification}, \emph{System Identification}, \emph{Filtering}, and other toolkits of using data to learn about the evolving state or parameters of a system. All of these have an underlying ontology, in posterior sampling as optimization over probability measures, stochastic (partial) differential equations, etc..

These applied computational mathematical physics fields present a bridge by which the objects of study truly become ``Being-for-itself'', from the perspective of Epistemology and Phenomenology. This instrumentality, that is the priority of \emph{usefulness} of the domain of knowledge to humans presents important considerations. One can witness, for instance, the dance of foundational mathematical theory and clinical practice, sometimes one leads the advance into a field of discovery, sometimes the other. 

Clinical intuition presents a curious underexplored realm of knowledge generation. Through many trials of embodied experience manipulating objects in a particular domain, it is commonly understood that expertise is cultivated, and this expertise provides insight into the system dynamics beyond the knowledge that is available in the field's canon of textbooks and research papers. This insight is clearly representative of special knowledge about the system and its dynamics of cause and effect, but often it may be incorrigibly intuitive, that is the clinician would not be able to write it within the grammar of the relevant scientific domain, or any language for that matter.




\subsection{Open Physical Systems} 
\subsubsection{Semantics of Emergent Causality}
The privilege and esteem Physics enjoys as a ``hard'' science is significantly assisted by the practical feasibility of establishing closed systems with near-perfect conditions for performing randomized experiments with all relevant quantities exactly specified. Standard laboratory experiments involve a device that, through careful incorporation of vacuum and sealing technology in the experimental apparatus, contains all of the potential mechanisms that generate energy or otherwise interact in substantive ways -- that is there are no unforeseen influences. Moreover, in most settings wherein scientific investigation in physics is performed, some well-developed theoretically guided sets of field and evolution equations are available, based on first principles understanding that is paradigmatic, and thus subject to limited correction by empirical experiment. 

By contrast, biological systems, including the physiological response of the human body to different ailments and medicines, do not satisfy either criteria. A human body is influenced by an innumerable set of factors throughout its lifetime, making a completely identified model representation impossible, as far as including all possible causal influences. Moreover, while the biomechanics of microcellular processes must, in principle, underlie the mechanisms associated with the aggregate effects of disease progression as well as remittance with the introduction of medicine, it is practically (that is, computationally) impossible to simulate a model that is of sufficient size to be interesting for the context. An entire first principles biochemical model of a cell is impossible to define in memory on any supercomputer, let alone an entire human body.

An \emph{emergent process} is a set of physical laws or patterns that appear at some macro scale relative to underlying mechanisms of micro scale components of the system. This is a regular theme throughout biology -- the Quantum dynamics such as with the Schr{\" o}dinger equation characterizing atoms' interactions as defined by their potential functions ultimately causes entire protein configurations, but it is impossible to model it as such computationally\footnote{in practice, coupled classical mechanics with density functional theory approximations of the quantum solutions are the standard workhorse for such purposes}. The metabolic, biomechanical, etc. systems defining animal physiology ultimately shapes behavior, which affects its survival and replication chances, which scales to ecological effects at the population level. Yet these scales are modeled in entirely different ways in practice. In such cases, the larger the scale and higher the layer of abstraction, the more imprecise and uncertain the model is. Whereas large scale biochemistry can still be simulated through approximations of the underlying first principles equations, more macro scales cannot. In e.g., ecology, equations are conjectured through a combination of intuition regarding influence and broad empirical observations. And in turn, the language of causality in ecology takes place in the layer of abstraction concerning entire populations of species interacting as predator-prey, resource competitors, etc. One can observe that in this field, generalized causal phenomena are judiciously claimed. The focus of scientific literature is on explanations and predictions of specific patterns and interactions, rather than the discovery of fundamental laws \cite{gaston2000pattern}. Chaos, or the ergodic properties of nonlinear evolution systems, is one of the few systematized scientific fields associated to emergence. Dynamic systems with a large maximum Lyapunov exponent but are globally bounded, and/or with a rich structure of singularity and bifurcation with respect to exogeneous parameters, give rise to an elegant theory of qualitative phase space analysis. The geometry of domains of attraction to equilibria as forming fractals is one of the most salient discoveries of nature's intrinsic beauty, captivating the imagination of both scientists and artists.

The procedure of performing research into scale-interactions across macro-meso-nano/micro is a challenging topic of great importance for much fundamental and interdisciplinary research. Whereas physics does involve emergence, a principled transition from particles to continuua underlies most theories -- see \cite{dyson2008fluid} for an example in numerical fluid dynamics. In regards to other large scale phenomena, e.g., in biology, such derivations are not available, and more heuristic and holistic thinking must be applied. See a general complex systems exposition in~\cite{svedin2005micro}. 

A key principle is that the \emph{reductionism} present in physics, that is a complete mathematical description of every relevant entity to precision, is too high a standard as far as understanding downstream scientific fields. This presents a challenge as far as the Epistemological underdeterminacy of any model, and thus less certainty in causal claims. In some cases, such as in ecology, one can perform imprecise analyses of broad trends that are informative. Furthermore, one can take care as far as generalizing any causal statement beyond some observed conditions. However, in medicine, accuracy and precision are paramount. Next we explore how this can be handled through aligning \emph{multiple} layers of abstraction simultaneously as evidence towards a particular causal claim.


\subsubsection{Biology and Medicine}
Causal Learning in Medicine and Health applications is an especially critical and active field of research in the present day. Indeed, the predictive quality and associated social welfare benefits from recognizing a \emph{true} cause versus an observed association, as far as potential for reliable pharmaceuticals, are clear. Thus, the approval or rejection of different medical procedures and pharmaceuticals, as informed by such procedures, should be grounded on conclusive evidence. 

While the literature on real data with conclusive causal discovery appears to be scant, there is considerable research activity as to developing contemporary Causal Machine Learning techniques for medical application, e.g.~\cite{sanchez2022causal,prosperi2020causal,richens2020improving,feuerriegel2024causal}. We can observe, however, that these works describe specific patient's progression at a given point in time, rather than more general causal claims of some input (environment, nutrition, etc.) resulting in the development of some affliction or disease. 

Let us study, then, the patterns of literature and communication in the relevant scientific fields that results in conclusions regarding cause and effect. In particular, we follow the specific language games associated with the wholescale research enterprise, which we observe is by necessity multi-disciplinary. 

We can observe that these fields have been largely isolated from the developments in the AI and ML communities. This is because the latter is characterized by the boom in deep learning, wherein big data and GPU computing clusters have facilitated the construction of highly overparametrized neural networks, which are capable of universal approximation (e.g.~\cite{shevchenko2022mean}). However, consider the domain of the most powerful statistical tool -- the randomized clinical trial. Due to clear ethical and logistical concerns, the sample sizes in such studies are in the order of 100s to 1000s, that is, many orders of magnitude less than the standard deep learning datasets with tens of millions of data samples. 


We now commence with describing a standard research protocol in the area of medical science, i.e., as an anthropological description of how causal inquiry proceeds successfully in this scientific domain.

The first moment a potential hypothesis regarding a medical fact is proposed is typically entirely observational, that is, some statistical relationship between a previously neutral quantity and a quantity of medical concern appears in the relevant data. Rather than being a targeted search initially, the hypothesis presents itself when otherwise ancillary to another medical study, or as part of a broader longitudinal data mining exercise to find unexplored new health and medical observations. Subsequently, domain experts in physiology, metabolism, biochemistry, etc. would consider the hypothesis, and either deny it as impossible, or propose a potential physically reasonable mechanism by which the observed, e.g., input, would have some effect on the body and in turn its health status that is in some sense related to, e.g., the problematic biomarker observed as correlating with the input in the data. 

Following this, more precise observational studies can be performed, now with targeted hypothesis testing specifically for the effect considered. The aforementioned contemporary tools in conditional independence structure become relevant after establishing that the assumptions, that is no latent confounders, can be at least compelling in particular settings. Finally, in the ideal setting, a randomized clinical trial wherein the variable of interest can be completely isolated, with all other variables marginally randomized, can be used to finally provide definite statistical credibility to the considered hypothesis.

Before finally considering the presence of cause and effect, however, additional research is performed at smaller scales, wherein systems of \emph{biophysics} become salient. Physically/Experimentally, this is performed with in vitro experimentation observing the effects of some injection in an isolated setting with the biological tissue of the same substrate as the setting of interest, e.g. a collection of red blood cells. Theoretically, this includes biophysical models of the application of known physical field theories to particular conditions present in the health phenomenon of interest, for example cardiovascular fluid dynamics~\cite{bergel2012cardiovascular}. 

Subsequently, additional observational and clinical trial data is collected. With a compendium of consistent results, a well constructed meta-analysis can result in collective agreement that the collective type 1 error is negligible, and with the consistent in vitro observations, the statistical evidence is clear that the observed correspondence is not random, but due to the presence of some consistent process unfolding, acting on the initial and exogenous conditions. This is followed by additional experiments, models, and an expansion of the set of domains, e.g. a discovery of the chemistry of the inputs causing certain intra-cellular biochemical reactions taking place at a higher rate, resulting in a sudden change in the downstream cascade, in credible simulated conditions of the milieu of the cellular environment.

This schematic description describes the grounded procedure of the medical research community associated with establishing a sound and robust consensus of cause and effect. Namely, it indicates that there is an iterative process between \emph{refined statistical precision} and \emph{credible mechanism}. That is, with each iteration, the statistical power of the tools is increased and the understanding of the low level mechanisms involved expands. 

The first iteration can be mechanism driven instead, of course. A practitioner from a lower level field, that is chemistry or molecular biology, discovers a new property of function of some component of a compound of interest. This prompts the flurry of activity described above. These iterations accumulate, until a \emph{preponderance of evidence} is established. 

It should be mentioned that a confident consensus, one that is reported as such to the general public and media, follows this under ideal circumstance. Indeed, the replication and reproducibility crisis in medicine~\cite{stupple2019reproducibility,blease2023replication} indicates the need for caution, and to carefully proceed with the entire scientific research program described above. With the presence of even fraudulant research publications \cite{kocyigit2023analysis}, overenthusiastic implementation of preliminary results in error can lead to elevated morbidity and mortality \cite{nato2022fraud}

To present a case study of scientific consensus leading to public consensus regarding medical
causation following decades of dedicated scientific practice, consider the uncontroversial claim that a regular habit of smoking tobacco causes long term health problems. Beyond the complexity of the human body and the long term physiological processes underlying human health, ethics and practicality make the Randomized Clinical Trials (that is, randomly assigning long term smoking and sobriety habits) impossible. Moreover, tobacco companies at the time actively sought to promote any possibility of doubt, requiring a high burden of proof for claims of health concerns to be validated. Thus, enough anecdotal observations leading to  initial survey data indicating statistical associations between lifetime smoking activity and poor health outcomes was not treated as definitive, but opened the scope for exploring the topics across multiple scientific domains and scales of time.

Biostatisticians worked intensively exploring intersectional permutations of possible confounding factors and health markers. In addition, however, biochemists identified carcinogenic and oxidative chemicals, conducted in vitro experiments on the effect of tar on lung tissue, and formulated an increasingly confident and complete narrative of arterial stress and plaque build-up. These all contributed to \emph{mechanism} in the physiological domains as to how smoking caused deleterious health outcomes. Over time, the collection of empirical and domain science evidence accumulated, and enough credibility was assigned to the proposition of deleterious health consequences to smoking that it affected public behavior of their own will as well as government policy. 

This is a clear case wherein mere statistical associations, even if well statistically powered, are simply inadequate in and of themselves for establishing true knowledge of cause and effect. Scientific endeavors that have proven to be the most useful and informative have incorporated multiple disciplines, with diligent proper methodology in each contributing to the component parts of the causal narrative. 

\subsection{Social Science}
Philosophers of Science have a consistent pattern of investigating the epistemology of the social sciences by systematically contrasting it to the natural sciences~\cite{rosenberg1988philosophy,risjord2022philosophy}. This consideration presents stark disadvantages to social science: the mathematical rigor, controlled experiments, and high standards of statistical significance that characterize the natural sciences are, for fundamental structural reasons, difficult if not impossible to incorporate into method and practice in the social sciences. The layer of emergence present in biology cascades, in the fields of psychology, sociology, economics, etc. into monumental intractability: dynamics in microbiology emerge as neurophysiology; dynamics in neurobiology emerge as patterns of thought; cognitive patterns emerge as behavior; agglomerations of behaviors emerge as social interaction and organization; changes in social interactions averaged across societies emerge as global sociological and economic patterns. 

Moreover, recent events have discredited the social science establishment and its authority in the domain - the omnipresence of a failure of replication and reproducibility has affected even hard core paradigmatic textbook theses of these fields. In this Section we present the central role of Hermeneutics as presenting a path forward for the embattled social sciences. This comprises of two central theses: 
\begin{enumerate}
    \item On the one hand, much greater humility should be practiced on the part of social scientists as far as the strength of certainty expressed with any claim of general mechanisms of cause and effect. As with biological open systems above, consistent findings across multiple scales and multiple disciplines must be established before any resolute conclusions. Fortunately, the field of hermeneutics and phenomenology, that is, descriptions of how concepts and events are understood and experienced, presents an additional field of inquiry, that is not present in the natural sciences.
    \item On the other hand, while the grand collection of established \emph{predictive} claims in the social sciences must be reduced by an order of magnitude, this does not reduce social sciences' utility to a commensurate degree. Their \emph{macro} utility is reduced -- significant caution must be undertaken as far as externalized consequences of predictive social science claims, meaning drastically limiting social science's influence on intrusive interventions in the lives of others by government policy or other influential actors. However, \emph{micro} utility can and should be maximized -- social science can provide guides and expansive catalogues of models, narratives, pictures, critiques, and guidance for appropriating narratives in literature, religion, mythology, etc. to assist individuals find \emph{their} truth as far as living a life that is full of joy, meaning, fulfillment, and overall flourishing in the mental space.
\end{enumerate}


However, in the consideration of people, narrative beyond mechanism becomes a potential methodology of understanding. That is, embodied human interpretation, by the practice and method of Hermeneutics with a primary focus on presentation of raw Phenomenology, is a unique tool among the social sciences for explaining a system. Thus, we argue that similarly to biology, wherein a combination of statistical patterns and description of domain science mechanism and multiple layers of abstraction can define a cause, in the social sciences this can come from a combination of statistical observations, evolutionary explanations of how a behavior arose, and symbolic meaning and stories surrounding the practice in both history and as experienced in survey experimental settings now. Thus, while the replication crisis presents caution as far as interpreting one-off results as definitive statements about how humans and societies function, by championing a standard of establishing \emph{preponderance of evidence}, that is simultaneous agreement across the domains described above, to establish new doctrines with respect to cause and effect in the social sciences.

In contrast to the natural sciences, the proximity to natural science practice as far as the use of mathematical models, textbooks of detailed specialized domain knowledge, as well as statistical precision and replication rates, increases with the scale of the domain of consideration, when we consider the central domains of psychology, anthropology, and economics. We list these below, briefly describing the salient characteristics of the scientific tribe and its grounding of a language game of causality. We present the standard methods of practice in the field, and provide considerations as to facilitating greater certainty and epistemic virtue as far as defining causal effects.

\paragraph{Psychology}
We begin this exposition on the nature of causation in psychology with two striking empirical observations:
\begin{enumerate}
    \item Psychology, especially social psychology, being the epicenter of the replication crisis, with the worst random sampled replication rates of the disciplines.
    \item As one of the most significant remaining findings that exhibits consistent harmony, psychotherapy generally speaking always seems to have a significant outcome in assisting recovery from mental disorders, with a reasonable effect size; however, the type and modality of the type of therapy (psychodynamic, CBT, etc.) did not present statistically discernible outcome differences. 
\end{enumerate}

Both of these points present strong evidence towards Gadamer's description of the Epistemology of Social Science. To the first point, there is a clear distinction from statistical studies in, e.g., physics, and social psychology. In the former, the system being studied in typically engineered to exist in a vacuum, and inverse problems are solved wherein the observation is set against an expected value of a particular quantity as determined by known physical equations of the system's operation. By contrast, a statistical model in psychology can be a basic, e.g., Generalized Linear Model, by which some demographic variable is expected to have a monotone effect on some outcome probability. No attempt at model representation, that is, presenting equations describing how this demographic variable could induce that outcome, is even made.

We can consider that the corresponding relative p-value standards of, e.g., particle physics (5$\sigma$) and social psychology ($p=0.05$), we can see that out of distribution generalization is going to be suspect in the latter. 

Note, however, that these are \emph{structural} problems. Fundamentally, the behavior of humans presents such significant uncertainty, that these two problems cannot be solved with more refined modeling and statistical methods. As mentioned above, human behavior is an irreducible system that cannot be faithfully characterized by a set of equations. Multiple layers of abstraction, significant open nature of the system, ubiquitous randomness and the ever-present possibility of praxis make definitive modeling intractable. Moreover, the task of carrying out experiments and surveys presents significant logistical challenges, and cannot realistically be developed for large sample sizes. This prevents the use of neural models in sharply defining the empirical observations of human choice. 

Clearly the behaviorist program of attempting to construct a methodology for the social science with isometries, as far as the language games of truth and causality to practice in the natural science, is fundamentally flawed.

Incidentally however, as described above, the second point presents a clue as to more effective practice. The second point underscores the \emph{Hermeneutic} truth of psychology. Any method of psychotherapy, CBT, Jungian, ACT, mindfulness, etc. appears to be equally successful, yet they all posit entirely distinct and contradictory models of the human mind and how dysfunction can occur. This suggests that on the one hand they are true, since they do have a measureable effect, yet on the other, they are not an ``objective'' truth, due to the contradictions. Hermeneutic truth, that is, compelling visceral narrative that inspires rightful and empowering action, is the proper Epistemology. This includes the language of cause itself, that is, a cause of a recovery to mental illness can correspond to a healing of an inner child or a cultivation of a mental reframing exercise (or both). 

Hermeneutic truth also provides the potential for the meaningful use of cultural knowledge. Anthropological ethnographic research into meaning, symbols, norms and practices provides interpretable insight as to human understanding. Next, we present the potential for incorporation of mechanistic explanations through evolution. This presents the potential for mathematical models as well as technical game theory, noting that population dynamics and evolution generate stability properties with correspondence to Nash equilibrium concepts~\cite{gintis2000game}. Finally, systems theory in cognitive science as well as neuroscience itself can provide additional description of the phenomenon of interest by describing physiological patterns of brain region activation that appear concomitantly with a certain cognition. We can observe from the language that there is often a shortcut with respect to the explanation, and the use of synonyms rather than the explicit word ``cause'' reflect the presence of multiple (in this case, mutation and evolution dynamics) intermediaries between the cause and the effect, for instance in standard phrases ``giraffes evolved long necks to reach higher treetops'' or ``the reason for the thick fur is the low temperatures of the winter cold''. 

An important observation is that the practice of psychology and research therein is intertwined with its \emph{clinical} practice. That is, unlike physics and even chemistry, there aren't long stacks of layers of abstraction that can describe a phenomenon, and extensive intellectual work in developing higher levels of abstraction completely disembodied from practice. Whereas, for example, particle physics concerns itself with fundamental ontologies, there is rarely a clean immediate correspondence between such a result and some engineering implementation. By contrast, as psychology concerns itself with people, there is an immediate necessity of utility to connect research results towards either indirect or direct communication to people at large or a subgroup thereof as to how they can assuage distress or obtain a psychic benefit. This fundamental distinction creates two consequences that make its class of valid methods fundamentally different than the natural sciences, rather than an inferior subset thereof: 1) Hermeneutic truth becomes valid in the sense that a viable standard for truth is simply a human understanding it, applying it, and believing it was of benefit, and 2) because only instrumentally useful knowledge is pursued, the standard of how complete and comprehensive theoretical substructures should be, before no marginal benefit is accrued, is much lower.

\paragraph{Anthropology and Sociology} present a range of curious considerations as far as the Epistemology of defining sound declarations of cause and effect. Such fields introduce the complexity of understanding broad patterns of collective behavior of groups of humans. These are typically associated with some tribe or geopolitical polity, or a subgroup within. We observe that a reductionist program becomes even more unrealistic, given that we must model interactions among large groups of individuals over a long time span, a computationally challenging problem even for well-identified engineering systems. We present several perspectives from well-respected practitioners in the field who can be said to have presented a definitive argument as to the presence of an interesting causal phenomenon, to elucidate the source of their credibility. We observe again the concomitant role of Phenomenological truth, as well as a more subtle definition of clinical practice and instrumentality. 

As a classical treatise, consider Max Weber's \emph{The Protestant Ethic and the Spirit of Capitalism}~\cite{weber2013protestant} presenting the thesis that: due to the main theological principles of Protestantism, this resulted in a culture that promoted conscientiousness and productivity and led to the unprecedented economic growth and development of European civilization. We can see how the emergence across the layers of abstraction works: the observed significantly higher GDP per capita (GDPpc) of the Protestant European nation states (at the time of the book's writing, although we can still see such states being overrepresented at the top of the GDPpc rankings) is associated with greater saving and investment as well as higher employment rates with longer hours, greater productivity and propensity towards entrepreneurial risk. This can be observed to correspond to behavior patterns as well as life principles the population lives by, as far as encouraging behaviors at the individual level that sum up to these macroeconomic principles. Finally, these behavior patterns and life principles can be seen to arise from the particularly Protestant reading of the literary content of the Bible. 

Let us contrast the comprehensive consensus that exists regarding the veracity of causal explanation present in Weber's seminal work with the highly contentious Frankfurt School and their associated Critical Theory. The Frankfurt School and subsequent offshoots was a very prominent social research activity, with both significant academic contributions to the field as well as popular reach of their works.  Founded in Weimar Germany in the early 1920s, they resettled in New York. They developed a heterodox Marxist tradition of social analysis. A foundational work in this direction is Horkheimer and Adorno's \emph{Dialectic of the Enlightenment}~\cite{horkheimer2002dialectic} broke from materialism in analyzing and showcasing how ideology can affect culture that contributes to the endurance of capitalism. This was further described as far as mechanisms within mass culture that subdue proletarian propensity for agitation by Adorno in \emph{The One Dimensional Man}. We finally mention \emph{The Authoritarian Personality} of Adorno that attempted to define a personality profile, caused by adverse childhood experiences, of a person prone to affiliation with oppressive militarist ideologies. We can see that much of the methodological criticism of their work~\cite{bottomore2002frankfurt} can be seen from the perspective of a lack of confirmed prediction accuracy, even to the point of theories fundamentally untestable. To this end, we can see that the school, while influential intellectually and politically, was light as far as descriptive statistics and quantitative study in general. Rather, the writing itself creates an aesthetic towards which compelling meaning is granted towards certain narratives of human motivation and mindset. With Marxism itself considered by other meta-narrative-identifying scholars to be either unfalsifiable or proven false, we can take care as far as interpreting claims that yield predictions that can be empirically tested. Rather, as Aesthetics, one can treat it as a set of ideas that ``taste'' in a potentially agreeable and compelling way, for some. These assist individuals in making sense of the world in a manner that is congruent to them, as far as a broader heterodox Marxist life philosophy and set of principles. Here we can see that by considering limitations as to the language games of their work, we can appreciate the insights while respecting limitations. At the same time, their effect is real -- a major theme in the Frankfurt school work is the Marxist concept of alienation, that is a laborer performing activities as a small component in an industrial machine, rather than the psychologically rewarding performance of generating a new useful object through will of action. As far as proper Epistemology, it is critical to treat such aesthetically compelling truths as the distinct language games they are relative to, for example, Weber's classic work. To the extent that individuals identify with the Frankfurt School and its canon, it can provide meaningful context for their individual choices in social and economic domains. As far as their dearth of predictive quality, it would be, ironically, overbearing authoritarianism to use them as principles by which to design government policy and political activity.

We conclude this Section with two recent and related examples by which consistency of causal explanations and descriptions of central mechanisms in society brings credibility to broad conclusions in social science. 

Let us turn to Bourdieu and his critique of the role of taste for class signalling~\cite{bourdieu1984distinction}. That is, rather than organic sets of preferences, or spontaneous cultural dynamics, tastes for consumer products and services reflect the wish, particularly of bourgeois and the upper class, to indicate to the surrounding world their superior status. We can observe that this sociological work is in harmony with an earlier treatise that studies the similar phenomenon from the perspective of economic discourse in \emph{Theory of the Leisure Class}, written by Thorstein Veblen~\cite{veblen2017theory}. 

Finally we can observe cross-disciplinary harmony in the following lines of work: 1) Bourdieu on Language and Cultural Power~\cite{bourdieu1991language} presents a description, from the point of view of cultural anthropology and sociology, of how language is used to propagate ideas and understandings that assist the well-being of the interlocutor's tribal and economic affiliations 2) \emph{Simulacra and Simulations}~\cite{baudrillard1994simulacra} describes the preeminence of symbolic capital, that is, economic investment and labor into purely abstract economic activity, such as marketing, media, academic work and other non-tangible production, becoming increasingly removed from any resemblance to grounded physical reality and 3) David Graeber's economic anthropology work~\cite{graeber2018bullshit} describing the phenomenon of a large and increasing component of the workforce performing clerical and symbolic work that adds little to nothing to economic productivity but expands and defines cultural norms and meaning within the corporate environment as well as public image. We can observe that each of these works touch on the same phenomenon, essentially, while taking different perspectives as far as its most salient manifestation as well as the field of inquiry and thus methodological toolkit for analysis. Note how the notion of symbolic capital is further in harmony with Bourdieu and Veblen's observation of consumption as signalling. 

Social science presents a clear demonstration of Wittgenstein's adage \emph{what can be said at all can be said clearly, and whereof one cannot speak, thereof one must be silent}. There exist profoundly useful insights in the social sciences, and the great tragedy is the nature and quality of meaning that has been ascribed to social scientific inquiry. 

\paragraph{Economics}
Economics treats an epiphenomenon, that is the agglomerated set of choices of buying and selling by large groups of people. It clearly falls into the realm of consideration of an irreducible open system. Yet the variety of methods that are used to engage Economics research is vast in language game semantics. Theory-guided practice includes: first principles axiomatic models, thought experiments, theory of games and cooperative/noncooperative behavior, econometrics, institutional analysis, historicism. Empirical methods include econometrics, behavioral economics, and experimental economics. Time and spatial scales include partial equilibrium models of instantaneous arbitrage-free liquid markets to long-term Kondratieff secular growth cycles spurred by a flurry of technological innovation. Epistemology of Economic thought deserves its own comprehensive work, and indeed the fact that the History of Economic Thought is studied in its own right indicates the pedagogical utility of considering the forms of methodology developed for this field over the years. 

Rather, to illustrate the appropriate framing of scientific language so as to promote harmonious synergistic insight, rather than unproductive intellectual tribal conflict, we turn to a less mature and rapidly evolving field within the social sciences: cognitive science.

\section{Anthropology and Psychology of Interdisciplinary Scientific Practice: Cognitive Science}
\subsection{Background}
“Cognitive science—it is not a discipline in its own right, but a multidisciplinary endeavor.” ~\cite{bechtel2017life}

Causality in cognitive science has historically centered around three principal questions: first, the relationship between the functions of the mind, second, the relationship between the mind and its putative substrates, and third, the relationship between these and the external world. For a comprehensive history (on issues beyond causality) of cognitive science, the reader should consult Bechtel, Abrahamsen, and Graham (1998) \cite{bechtel2017life}. Thus, epistemology in the cognitive sciences and the exercise of finding fundamental cause-effect phenomena presents a fertile example of multidisciplinary research. It deals simultaneously with phenomenology and cognition (the forms and functions of the mind) while attempting to reconcile these with models of the substrate producing said mind. Due to the myriad of 
motivations and approaches to these central problems, the language game semantics for causative frameworks include complex systems, computer science, psychology, biology, chemistry, physics, and mathematics. Cognitive science, due to its focus on both substrate and phenomenology, presents intriguing points where the nature of causality itself begins to enter points of indeterminacy. This is evident in the many debates surrounding the relationship between phenomenology and neural correlates, a subject that remains far from settled. Yet a paradigm for normal science has been established, with laudably quick adoption of standards in the field for reporting results and making inferences, e.g.~\cite{poldrack2008guidelines}. A second aspect of complexity in the cognitive sciences arises from its intersection with computer science. Computer science studies best practices when it comes to computation for reasoning about the world, and a biologically adaptive brain would reasonably compute more accurately and efficiently than otherwise. Thus the field of artificial intelligence both informs and is informed by existing models of consciousness and cognition. This extends beyond philosophical speculation into the realm of engineering application, producing fascinating work in which objects of phenomenology and the general problem of life are beginning to be formalized in novel ways (see contemporary accounts of trust, empowerment, utility, affect, curiosity, etc.). However, it is important to note that while neuronal dynamics have been an influential source of algorithms, practitioners must generally take care in making inferences between the two given the observed pervasiveness of biases and heuristic reasoning in human cognition that one would not typically want to imbue within a computing machine. Similar care must be taken when applying models derived from technology to understanding the brain, e.g. early thermodynamic descriptions of the unconscious, or contemporary abstractions of neural networks then applied to the still-unfolding complexity of brain function.

Contemporary approaches to understanding causal mechanisms in cognitive science have become increasingly sophisticated and multifaceted. The predictive processing framework~\cite{clark2013,hohwy2013} has proposed that the brain operates as a prediction machine, with reasoning flowing both ``top-down'' and ``bottom-up'' through hierarchical generative models. Embodied cognition approaches~\cite{varela1991} have emphasized how cognition extends beyond the brain to include the body and environment. Dynamic systems theory~\cite{thelen1994} has provided mathematical tools for understanding non-linear  interactions in cognitive development and function, while complexity theory has offered new ways of understanding emergence and self-organization in cognitive systems. As such, cognitive science is one of the most multifaceted as far as its incorporation of a variety of language games, because even at its most simplistic, it is concerned with (1) the functions of the mind, which is amenable to many different approaches due to the need for analogy (whether poetic, literary, arithmetic, etc.), (2) the identity, nature and function of the substrate of the mind (whether in neural tissue, the body, combinations thereof, or different scales of the substrates ranging from individual neurons to ensembles to whole brain areas), and even today, the persistence of dualistic theories that posit extra-material substrates, (3) the functional relationship between various conceptions of the substrate and aspects of the mind as well as subjective experience—that is, the first-person perception of the mind—as well as (4) the problem of perception, namely, the form and nature of the correspondence between the objective world (a concept that carries its own baggage) and the contents and substrate of the mind.

The sheer number of competing accounts regarding each of these sectors and of the relationships between them, each of which is challenging in its own right, leaves us with the problem of a daunting permutational complexity that is frequently difficult to reduce due to the bridging of scales and disciplines that are frequently incommensurable.

\subsection{Literature}

Here we present a taxonomy of a list of a number of prominent (although not exhaustive) subdomains within cognitive science. Each of them represent the development of a theoretical strategy, method of experimental practice, and standards of statistical testing associated with a distinct tribal language game associated with a particular Science. 

\paragraph{Cognitive Psychology}
\begin{itemize}
\item \emph{Purview}: Studies mental processes including attention, memory, problem-solving, and decision-making through controlled experiments and theoretical modeling. Experimental manipulation of stimuli/conditions yielding measurable changes in behavioral responses and task performance provides much of the published literature.
\item \emph{Cause Semantic}: Terse description of structure and function of different abstracted modules of cognition, such as memory, sight, etc.. 
\item \emph{Quantitative}: Primarily Empirical/Statistical with some Theoretical modeling
\item \emph{Narrative}: Limited; primarily uses technical scientific descriptions
\end{itemize}

\paragraph{Neurocomputing}
\begin{itemize}
\item \emph{Purview}: Develops computational systems and algorithms that emulate cognitive processes, focusing on learning, reasoning, and problem-solving capabilities. Based on spin-glass theory, this line of work provided the first motivation for the development of neural networks in the early days of AI.
\item \emph{Cause Semantic}: Mathematical/algorithmic formulations, through their physical energy function properties, result in specific computational outputs
\item \emph{Quantitative}: Theoretical+Simulation
\item \emph{Narrative}: Limited; uses technical and mathematical descriptions
\end{itemize}

\paragraph{Linguistics and Psycholinguistics}
\begin{itemize}
\item \emph{Purview}: Examines language structure, processing, acquisition, and production through both formal analysis and experimental methods
\item \emph{Cause Semantic}: Sensible conceptual model together with empirical validation of linguistic structure and semantics correspondence to emotional or attention affect of the interpreter. Field is mostly descriptive rather than causal mechanism focused, however, with some standout exceptions proving the rule (e.g. Chomsky).
\item \emph{Quantitative}: Both Empirical/Statistical and Theoretical
\item \emph{Narrative}: As narrative is often the target of the research itself, method can be recursive and self-referential. There is a semantics of technical description and linguistic examples/narratives
\end{itemize}

\paragraph{Philosophy of Mind}
\begin{itemize}
\item \emph{Purview}: Epistemology is the philosophical study of the mind generally, with a specific focus on belief, knowledge, and certainty.
\item \emph{Cause Semantic}: Meta, e.g., this paper discusses Causation and can be considered Philosophy of Mind.
\item \emph{Quantitative}: Limited with rare exceptions (e.g. Badiou)
\item \emph{Narrative}: Extensively employs a variety of narrative language games, with a particular prominence to thought experiments
\end{itemize}

\paragraph{Cultural Anthropology of Cognition}

\begin{itemize}
\item \emph{Purview}: Studies how cultural contexts shape cognitive processes and identifies universal cognitive patterns across cultures
\item \emph{Cause Semantic}: Cultural practices and beliefs shape cognitive processes and behaviors
\item \emph{Quantitative}: Extensively Empirical/Statistical with some Theoretical Modeling
\item \emph{Narrative}: The structure of meaning, in particular ethnographic description and translation across peoples is central to the field.
\end{itemize}

\paragraph{Cognitive Neuroscience}

\begin{itemize}
\item \emph{Purview}: Maps cognitive functions to neural activity and brain structures using imaging and other physiological measures
\item \emph{Cause Semantic}: Consistent correspondence of neural activity being associated with cognitive functions and behaviors indicates the latter is an epiphenomenon of the former. 
\item \emph{Quantitative}: Empirical/Statistical with heavy reliance on imaging data and the (exploding in quality) Computer Vision ML
\item \emph{Narrative}: Limited; primarily observational with parallel descriptions of brain physiology and general cognition modules and their main functions.
\end{itemize}

\paragraph{Developmental Psychology}

\begin{itemize}
\item \emph{Purview}: Studies the emergence and transformation of cognitive abilities across the human lifespan. Presents a description of how patterns of cognition change with age as well as standard and possible uncommon life events, with a focus on the rapid brain development undergone in childhood. 
\item \emph{Cause Semantic}: Mostly descriptive, however the usually implicit cause is activation of genes, sometimes mediated by environmental influence, yielding change in biochemistry that leads to changes in cognition.
\item \emph{Quantitative:} Primarily Empirical/Statistical with occasional Theoretical modeling
\item \emph{Narrative}: Fundamentally historicist: aims to present an interpretable narrative of the correspondence of life history and change in cognition patterns. Thus, methodologically the science uses both technical methods and illustrative case studies.
\end{itemize}

\paragraph{Evolutionary Cognitive Psychology}
\begin{itemize}
\item \emph{Purview}: Studies the ecological pressures, and thus the utility for different cognitive processes, as well as what cognitive processes could be conjectured to appear.
\item \emph{Cause Semantic}: Selection pressures in an ecological sense
\item \emph{Quantitative}: Theoretical with Rich Simulation
\item \emph{Narrative}: Uses historical narrative descriptions defining events in a lifetime that can affect an early hominid’s survival of replication chances in a particular environment.
\end{itemize}

\subsection{Challenge to Method or Rich Fertile Ground?}
Remark on the self-referential and recursive nature of this section embedded in this paper, that is, a discussion on the Anthropology of Cognitive Science, including the Philosophy of Mind, in a paper presenting Philosophical understanding and framework of knowing (about causality specifically). We proceed to remark that there is a correspondence across language games that exhibited meta-narrative concepts of insurmountable challenge, that is better interpreted through Ordinary Language philosophy, as, rather, opportunity. 

Cognitive science is a unique scientific field in that, with relatively equal emphasis, it studies both the substrate and functions of the mind from the exterior, as well as the experience of that substrate - itself having a mind. Or, as some have noted, psychology is the study of physics in the first person. This dual perspective encompasses what has been called the ``hard problem'' of consciousness: how and why subjective, phenomenal experience (qualia) emerges from physical processes, and whether this explanatory gap can ever be fully bridged. 

We contend that the more proper framework is ``exciting opportunity'' rather than ``hard problem''. In doing so, we move to the philosophy of Mind, wherein we observe this ``hard problem'' is associated with, in the Anglo-American Analytic tradition, realist and anti-realist debates on qualia using various thought experiments often involving a brain in a vat. 

Let us now perform Ordinary Language Semiotic Deconstruction, as is done throughout Wittgenstein's work. This is the method by which a normative philosophical quandary is reduced to its constituent phrases and interpreted precisely as ordinary language, that is, how those phrases are used in ordinary life, lifting a mirage of profundity to reveal a question that is either vacuous or with a disappointingly trivial answer. Let us engage the Analytic Philosophy of Mind Tribal Debate Language Game space: consider the question of whether there is ``free will'': we must first understand what the ordinary language of the term means. When one uses the term ``free'' it is at least implied, and usually stated outright, that it is free \emph{with respect to something}. Specifically, ``free'' is the capacity to act independent of some explicit or implicit restraint, such as a physical obstacle or restraint (``was finally free from the bondage''), a circumstance like imprisonment (``freed from prison''), or as referring to a large set of possible actions that are proscribed by an oppressive government (``we're not free to speak our minds here''). In a question of the existence of free will, what would a will be free with respect to? The neurochemistry of the brain? That would certainly be physically impossible. Nor would it be preferable, as an individual would rather their will be decided within their brain rather than some unknown external force or individual. Free with respect to past influences and genetics? One naturally turns to an instrumental notion as far as psychotherapy to reduce traumatic influence of childhood trauma and contemporary patterns of behavior. But this becomes closer to freedom in an Existentialist (i.e. Continental, rather than Ango-Analytical) sense, that is the Ethic of living life true to oneself, thus not what is intended by the Analytical vat brain linguistic swordsmithing. We can see that the source of the confusion is that neuroscience of the brain and psychological observations of genetic and childhood influence are straw-manned here, implicitly, as an \emph{only} cause. But it is an only cause only under naive understanding that cause is confined to one level of abstraction/scientific language game. In the case of human phenomena, we see that it is not. Neuroscience and developmental psychology do not invalidate the information of other language games, like those of behavioral psychology or even praxis. 

As can be inferred from both \emph{Philosophical Investigations}~\cite{wittgenstein2009philosophical} and \emph{Culture and Value}~\cite{johannessen1995culture}, rather than being an insurmountable philosophical question as to the correspondence of qualia to neurochemistry, that is the ``hard problem'' which is assumed to have a singular definitive answer, what the hollowness of the debate indicates is the impulse and desire for \emph{awe}. That is, the fact that there is a rich inner life, and at the same time we can perform brain surgery and brain imagery and there is some correspondence of cellular models and physical properties of brain substrates to this inner life, is, and rather should be, awe-inspiring. This fundamentally stark distinction of form, simultaneous with intimate correspondence of ontology, is absolutely fascinating, and thus, rather than a singular ``hard'' problem, requires a corresponding expansion of vision as to how to answer a problem. To this end, the classic metaphor of a group of individuals seeking to describe an elephant in the dark while located at various places around the elephant applies. None of them can describe the elephant in its entirety, but each of them can describe a portion of the elephant, and the knowledge gain of their communication to each other is additive. The ``hard problem'' is actually an opportunity for a sizable number of fields, as indeed the 8 above are still missing several major ones and a number of small less developed fields, to be put to work to each contribute to the understanding of the qualia to brain tissue activity correspondence. 

We next present some particularly promising themes for cross-domain harmonization, dialogue, and meta-theoretical subsumption. 

\subsection{Prominence of Storytelling and Role of Hermeneutics}

Many studies demonstrate that the mind appears remarkably adept at ordering perceived events in time in a manner that aligns with affect, intent, and perceived agency. It functions more as a storyteller maximizing meaning than as an objective recorder attempting to maximize truth. This has been attested to by classical cognitive psychology and millennia of contemplative traditions.

This storytelling nature of mind interacts complexly with prediction and predictive processing: our experience of the present moment is shaped by both our predictions about what will happen and our interpretations of what has happened. The relationship between attention, conscious awareness, and different timescales of neural processing further complicates this picture, as different aspects of experience appear to be processed at different rates and integrated in ways we are only beginning to understand.

A number of findings in Behavioral Experimental Economics found the persistence of \emph{cognitive biases}~\cite{caverni1990cognitive} in human cognition that predictably lead to erroneous models of the world. If we consider that this must be a feature, rather than a bug, as due to some evolutionary adaptation, and indeed it is currently believed that rather than accurate cognition about the world, storytelling was more advantageous for early hominids due to its potential to facilitate cooperation~\cite{coen2019storytelling}. This is reflected in the large set of findings indicating that the function and purpose of language was predominantly social, that is to persuade, cajole, shame, etc. rather than to present accurate facts~\cite{enfield2024language}. 

Consider how the semantics of literature, and informed engagement as far as understanding the themes and aesthetics of the book, and providing a description of psychological states and personality profiles of the individuals, could be performed in similar language by a literary connoisseur, a critic of culture, and a psychologist.  Thus the work of literature is meaningful, and can be trusted to have mutually understandable meanings, to all three readers. Points of emphasis would vary, as far as wordplay, lessons of virtue, and psychoanalytic insight, but the interpretation of the story itself is ground to all of them the same. 

We can see how this presents a unique opportunity for social science to engage with subjective meaning and interpretation. The radical subjectivity of private meaning is not a caution tape, rather it is a unique form of insight into understanding some event or phenomenon. Moreover, the meaning of a story to individuals and tribes provides a form of intellectual inquiry without the years of devotion towards cultivating expertise in a more technical science. This presents the opportunity of distinction of the mathematical natural sciences versus the social sciences as one of tradeoffs rather than indisputable hierarchy. In the first, the presence of a cause requires a dynamic equation sufficiently validated by experiment, which is challenging exercise of significant cost in both human capital development and experimental apparatus. In the second, meaningful insight can derived from parsing literature and other cultural language, which is significantly less capital intensive, and with a much shallower learning curve.

\subsection{Theoretical and Empirical Multiscale Substrate Analysis}

Because our models of mental function continue to evolve, there is no clear endpoint or guarantee of a final version in our accounting or framing of the signals deriving from the mind's substrate. This presents a fundamental challenge to definitively adjudicate models, but rather suggests methodological heterogeneity. That is, in the literature there is often debate as to one representation of substrate being more useful to perform inquiry with than others, however, while some sporty rivalry can motivate gusto science, beyond that the focus should be on the rich variety of explanations, as far as the physiology of the elephant, and how they inform, rather than potentially debunk, each other.

The substrate of focus ranges from sub-components of the brain at one extreme, to latent factor graph constructions in the extended mind/extended brain hypothesis. There is also a notably prolific field of computational and information-theoretic approaches, each presenting a rich informative set of functional and structural explanations. 

Even the data can be cross-purposed, and the appropriate observation technique and statistical method to perform empirical research on cognition is not self-evident given the topic of interest. One can relate events or treat the data as an overlay of memory, prediction, and present-focused processing and attempt to identify the processes. Scientific interest in the nature of ``the present moment'' has yielded a new field, neurophenomenology, which attempts to bridge first-person and third-person perspectives by training subjects to provide more precise phenomenological reports while simultaneously collecting objective measures.

We remark on Marr's influential framework of implementation, algorithm, and computational levels~(see e.g.~\cite{peebles2015thirty}), as far as a prominent framework for integrating multiple layers of abstraction. This early detailed development of theoretical studies of emergence, earlier and more mature than other fields, is indicative of the rich potential for multiple explanations for cognition. 

Methodological questions on considering emergence also arise at the empirical, signal processing level.
The work of Logothetis and colleagues ~\cite{logothetis2003underpinnings} exemplifies the challenge of reconciling signals derived from vastly different scales of brain function within a single well-defined statistical model. 

Consider the daunting complexity of simply relating functional magnetic resonance imaging (exemplified by the BOLD signal) to the single-neuron recordings of electrophysiology. The first is an aggregate signal derived from differences in the magnetic properties of oxygenated and non-oxygenated hemoglobin in cubic millimeters of neural tissue. As vast populations of neurons change their firing patterns, layers of complex, nonlinear processes translate their metabolic needs into increases in circulation, changes that can occur on the order of 5-10 seconds and affect 100's of 1000's of neurons, each of which is directly connected to 1000's of other neurons, and indirectly affects 1000's more, through mechanisms that are still being elucidated. In contrast, electrophysiology can only track the electrical activity of numbers of cells in the hundreds, or thousands at the bleeding edge. Between these extremes lie other crucial measurement scales: calcium imaging offering intermediate resolution, EEG/MEG recordings providing different temporal/spatial tradeoffs, voltage-sensitive dye imaging, and emerging high-density multi-electrode arrays capable of recording from thousands of neurons simultaneously. Each of these methods captures different aspects of neural dynamics, from millisecond-scale events to slower brain rhythms operating at multiple frequencies, simultaneously.

On the one hand, observe the interplay with statistical signal processing, for which the development of statistically powered multiscale integration models is a challenging exercise in applied probability~\cite{shen2012multiscale}. On the other, there are fast neural events and slow hemodynamic responses that correspond to these. A self-evident research program of studying emergence is the obtaining of qualitative estimates from data, e.g., the degrees of freedom at each scale as far as mutual interaction, and noise representation from first principles, e.g. fractional diffusion due to weak interactions of certain chemical bonds.

\subsection{Cognitive Science to Artificial Intelligence as Physics to Engineering?}

The development of artificial intelligence as an automated reasoner presented new opportunities for informative methodology. The structure of the brain, represented through interconnections of neurons with activity of connections modeled with statistical mechanics of spin glasses, presented a number of seminal models in the history of neural networks, such as Hopfield networks~\cite{hecht1989neurocomputing}. Extending and fine-tuning these, and aligning them to serial computing control flow yielded the feedforwark networks that formed the foundations of the, e.g., convolutional and attention network models today. 

However, beyond this foundation, AI quickly became superior to human cognition as far as arithmetic, then increasingly more complex tasks up to, e.g. winning at chess and go. Meanwhile, neural architectures have largely eschewed Hopfield models in favor of deep compositions of Rectified Linear Units (for example). The inspiration of the brain for computing has not played a major role in algorithm development in recent years. Moreover, the discovery and increasing awareness of cognitive biases presents the obvious dead end of seeking greater precision and accuracy in understanding through understanding 
human scientific cognition. 

Recently, interest and research activity has arisen as far as modeling emotional and social patterns of cognition in NN settings. Hypothetical simulations of explicitly programming behavior associated with decisions and beliefs that are derived from what would be the domain of biology, e.g. hormones and vasoconstriction. This includes neural models of emotional systems, homeostatic regulation, and inductive biases pertinent to the persisting in the physical world.

Still one can consider if more is to be gained from a better representation of the development of human cognition. That is, a standard ``scaling perspective'' posits that increasing model capacity, training data, and architectural sophistication will lead to emergent general intelligence through the ability to solve increasingly complex problems in increasingly flexible ways.
The embodied cognition approach suggests that intelligence and mind can only emerge in a system tasked with maintaining itself in the face of radical uncertainties posed by the physical world and the iterative evolution of agency in the context of other agents. This view emphasizes the essential role of physical interaction, sensorimotor contingencies, and bodily regulation in developing intelligence. Developmental robotics extends this perspective, suggesting that intelligence must emerge through stages of physical and social development, similar to human ontogeny.

Hybrid approaches attempt to bridge these perspectives, suggesting that embodied experiences might be combined with large-scale learning systems, or that symbolic reasoning might emerge from subsymbolic processing given the right architectural constraints. These approaches often emphasize the role of social interaction and cultural learning, suggesting that neither pure embodiment nor pure computation is sufficient. We see here already the proliferation of multiple explanations of the same phenomenon, multiple empirical and theoretical tools, and even systematic attempts to integrate them. 

\section{Primary Conclusions}

\subsection{``Cause'' has Distinct Forms across Scientific Language}
In exercising the Ordinary Language method of Philosophy as attempting to correct ``the bewitchment of our intelligence by language'', we considered the comprehensive set of circumstances wherein the term is used, and noted its appearance in every one of the natural and social sciences. At the same time, the actual form of a statement that asserted that one phenomenon caused another starkly varied, e.g. from differential equations to complex narratives. This suggests that the meaning of cause is one that is functional within a scientific language game, and we could also see that it plays a prominent role as to the priorities in a reigning scientific paradigm. 

We can observe, in how the term is used across the sciences, that ``cause'' describes the primary mechanism by which things function in the science. For instance, in physics, there are four primary forces, and it is standard to say ``gravity caused this object to move'', ``an electric field caused the capacitor to increase in voltage'', etc. In chemistry there is a hierarchy of bond types of increasing strength, and in, e.g., socioeconomics, cause is often used in the set of changes in exogenous quantities to the system after it comes to a new equilibrium, for instance ``introduction of this technology increased marginal productivity which resulted in lower marginal costs, resulting in a lower price, greater quantity sold, and higher wages, and this in turn caused immigration to increase''. Thus it can be seen that causation of this varied form is an essential component of the primary foundation of that scientific discipline. 

Consider the example of the following distinct set of causal statements:
\begin{enumerate}
    \item \emph{Physics} the Lagrangian of a falling object can be solved to reveal how the potential energy decreases and the kinetic increases. $L(q,\dot{q})=T(\dot{q})+V(q)$
    \item \emph{Chemistry} Adding Sulfur to Iron causes them to form into iron sulfide Fe+S$\to$ FeS
    \item \emph{Bio-Medicine} Trans fats consumption causes low density lipoprotein concentration to increase and high density lipoprotein concentration to decrease, causing blood vessel and heart damage long term
    \item \emph{Psychology} Witnessing at intense episode of violence can cause difficulties sleeping and hypervigilance in people
    \item \emph{Economics} The presence of asymmetric information between the buyer and seller causes market failure, that is a market outcome that is suboptimal as far as social welfare
    \item \emph{Anthropology} Mythological stories describing the history of the tribe's settlement of the highlands causes a sense of unity of purpose among the population.
\end{enumerate}

We have reviewed that establishing a cause with certainty is a challenging question for Epistemology for most of these scientific domains. Physics related disciplines enjoy several unique features that do not appear in the biological and social sciences. By contrast, for open complex systems, it is critical to demonstrate an aligning set of causal dynamics at multiple layers of abstraction, that is, preponderance of evidence hinges on testing and investigating the phenomenon of interest in multiple domains of inquiry.

\subsection{Critique of Causal Machine Learning}
The fundamental matter of concern as far as the use of terminology to the effect of ``Causal Learning'' and ``Causal Inference'' is not that there is no validity at all to the statistical methods involved as far as being relevant to the understanding of causes. Rather, the bewitchment comes because the computation of a particular DAG and set of conditional independence hypothesis tests do not conclusively present evidence for a causal relationship between variables (unlike, e.g. as implied explicitly in~\cite{eberhardt2009introduction}). And, fundamentally, nor can it. Since ``cause'' is anthropologically contingent, as far as a scientific language game, a causal inference can only be made in the logic of that scientific discipline, and establishment of causes in the accepted practice of establishing positive new results on the mechanisms in the field.

The risk then is overconfidence. To quote Wittgenstein: ``What can be said at all can be said clearly, and whereof one cannot speak, thereof one must be silent.''\cite{wittgenstein2023tractatus}. The phrase ``this set of numerical results imply that A is a cause for B'' is not correct, rather, they may suggest evidence towards that possibility. As it is, Causal Learning is in the language game of learning techniques, which is computational mathematics. 

Yet, we can still perform an inverse meaning calculation to attempt to establish a synergistic relationship between Statistical Causal Learning and Causal Narratives, as far as noting some intuitive notions that appear from the statistical representation. We observe that conditional independence is akin to the notion of randomization in a Clinical Trial: by sampling independently from a population you highlight a particular treatment. From both the form of Bayesian Dynamic Networks as well as some CP-Logics, the notion of temporal ordering is also significant, that is: causes precede effects. Notions of mediation, wherein A's effect on C depends on B, have an interpretable narrative associated with it. And so we can cast \emph{Causal Learning} as, more accurately, a \emph{Statistical Model with Interpretable Representations of Object Relationships}. While not as bold as claiming to conclusively define causes, this is still a prominently useful research activity. 

\subsection{Epistemic Virtue and Causal Reasoning}
Consider now the perspective of Epistemic Virtue~\cite{montmarquet1987epistemic}, wherein the emphasis is not on an investigation as to what it means to know something, etc., but rather the real practice of pursuing some intellectual inquiry with full clarity of method and practice, and due diligence to make an effort to acknowledge all relevant evidence and pursue understanding in good faith. The conclusions from this monograph point to a virtue entailing taking great care when considering proclaiming a novel attribution of cause and effect within any scientific domain. What would previously constitute a purely statistical identification of a causal phenomenon can no longer be understood as such, but as one result that suggests a particular association and direction of influence that must be thought of in the context of its juxtaposition with all the related results either in that domain or the same topic of objects and relationships of interest but across different domains and layers of abstraction or spacetime scales. One can consider the fact that one reason for overconfident research reporting is the bias towards publishing a positive outcome, and a bias towards claiming to know more about the world, thus making the claim of causal determination incentivized. We can consider instead that a publication would present the novel statistical observation as an indication of a functional relationship of a certain form being more likely than before, and then proceed to look through the literature and present harmony or disharmony explained by the various domains, and how the multilevel causal explanation is enriched or degraded by the new discovery.

\subsubsection{Role of Forward Quantitative Methods: Theoretical Models}

\paragraph{Badiou and Mathematical Ontologies}
Consider the perspective of Badiou~\cite{badiou2007being}, that mathematical formalisms are the only ontologies. When we posit an equation system to govern a particular phenomenon, if prediction error is low, near machine precision, the equation system \emph{is} the phenomenon. Since causation is a fundamental principle of ontology, then, the forcing terms in the evolution equation are the causes. When, however, there is a significant amount of uncertainty, that is, commensurate with the scale of the terms in the equation, then the event has not been fully accounted for by the aforementioned equation system, there is simply more to the story, i.e., additional ontology that can be understood as perturbation expansions of, or unknown unidentified nonlinear terms in the equation system. 

To give a concrete example, when we define, e.g., a system of differential equations governing the dynamics of a Hamiltonian system,
\[
\dot{q}=F(q,p),\,\dot{p}=G(q,p)
\]
we can immediately ascertain that $q$ changes based on the force $F(q,p)$ applied, depending on $q$ and $p$, etc. By defining these to be, e.g., the position and velocity of an object, one can even visualize, e.g., an object flying up and coming down, or a pendulum. We can see this principle that the cause of some change is defined by the right hand side in a differential equation or evolution PDE system, such as,
\[
\frac{\partial Q}{\partial t} = -a
\frac{\partial^2 Q}{\partial x^2}+F(P)\cdot \nabla Q+cQ
\]
where a population of a species diffuses across the environment at a rate $a$ while being subject to predators $F(P)$ and self-propagation of rate $c$. Again one can visualize, in e.g. a plane, an amoeba-like object translucent according to the value of $Q(x,\bar{t})$ at current time $t$ and the amoeba's size being the support, that is the value of $x$ for which $Q$ is nonzero. The presence of a predator appears as a disappearance of color in sections of the amoeba. 

Fitting these equations to data reveals their precision. Typically in physics and engineering, wherein upwards of $99\%$ of the exact function is known with only mild parametric uncertainty, when the evolution equation is fitted to data, decreases an already low numerical solution error correspondence to data to machine precision. In this case, most of the model dynamics has been identified, and so the causal mechanisms are directly visible and understood in the equation systems. These in turn can be simulated to show the effect on the trajectory of the system. 

By contrast, consider the ecological modeling, wherein the equations are not taken from a comprehensive known theoretical structure but arise out of a combination of intuition and empirical observations. In this case there is significant bias and variance of the residuals of the model fit to the data, forecast forward in time, and prediction across different instances, in slightly different circumstances, of the same dynamics. As a result, unlike in the previous example, we are simply not done. This is a case wherein refinement of knowledge can proceed iteratively, with out-of-distribution error as an indication to continue to fill out the map, or the shape of the elephant, by introducing additional elements to the model to account for more phenomena and refine understanding.


\paragraph{Complex Systems} is a theoretical framework grounded in reasoning of ergodic properties of a large number of objects with irreducible and often multi-domain interacting coupled components. While quantitative forecasts of such processes may be unrealistic, ergodicity presents a qualitative depiction that is less sensitive to the complexity of the functional form of the forces in the system. Thus Complex Systems provides techniques for ascertaining qualitative asymptotic properties of physical phenomena as well as reasoning from first principles and simulations to test and illustrate the reasoning. 

Complex Systems is grounded in the deep mathematical field of ergodic theory. Ergodicity is the study of the long run limiting properties of some dynamic phenomenon, including the attractors as well as the domain of attraction. This can be the fixed points and their stability in Dynamic Systems, or a more complex topology with evolution PDEs~\cite{ladyzhenskaya2022attractors}. This would correspond to stationary distributions of stochastic processes, e.g. the rich theory of Markovian processes in the seminal~\cite{meyn2012markov}. Of course, there is a rich variety of long run behavior, with alternatives like quasi-ergodicity, periodicity, and strange attractors with chaotic systems, all being phenomena that appear as qualitative long run states. Even systems with no asymptotic stationarity, that are dissipative or have other secular behavior, can be studied from the standpoint of ergodic theory. These give either robust stable systems, or a rich fractal geometry of possibilities, depending on the circumstances.

Because of the presence of recognizable global qualitative phenomena whose type persists with the input, it is possible to develop a form of reasoning about systems. Specifically, given a system in a current online ergodic state, one can consider exogenous changes to the inputs of the system and how that affects force terms. This would in turn shift the equilibrium point or distribution. This presents an opportunity to engage in reasoning at a lifted level of abstraction as far as causality in systems theory. Furthermore, by defining a complex system with sufficient formalism to characterize the necessary elements of the components and their interaction, one can perform simulations using synthetic data to learn about such systems. For instance, one can say that a system truism is that if a certain behavior is incentivized, then the total portion of the population engaging in it will increase. 

This has been standard in Economics, implicitly or explibitly. A supply demand curve represents, for example: a sudden increase in the demand shifts the demand curve to the right to a new equilibrium where it intersects the supply curve at a higher price. This corresponds to more people buying a product than before, inventories at shops running short quicker than before, and then either due to a logistics operator noticing or the management noticing, deciding to increase the stock of that product from the distributor. The distributor agrees while also suggesting to increase the price. This is a process that takes time but has a reasonable mechanistic explanation. Complex Systems Theory has been applied to History~\cite{turchin2011toward,tainter1988collapse}, understanding of the brain~\cite{telesford2011brain,badcock2019hierarchically,singer2009brain} and many other phenomena.

\subsubsection{Role of Inverse Quantitative Methods: Statistics and Learning}

Consider the matter of estimating or learning phenomena. We can observe that there is an important role of the forward system as defining the statistical model representation. For instance one can develop a very precise model of a physical system governed by ODEs and PDEs and solving inverse problems to find the parameters. Formally,
\[
\begin{array}{rl}
\min\limits_{p\in\mathbf{P},u,z} & \sum\limits_{n=1}^N\sum\limits_{s=1}^T \|\hat{o}_{n,s}-O(u(\tau(s)) \|^2\\
\text{subject to } & \dot{u}(t)= f(u(t),z(t);p)
\end{array}
\]
where $p$ is the unknown parameter, $z(t)$ is an input, and $u(t)$ is the state of interest. The function $O$ is the observable, and $\hat{o}_{n,s}$ are the observations for trial $n$ at time $s$. We can consider that the functional form of the system $f$ is known, and the dynamics are to be fit to an unknown low dimensional parameter $p$ within some set of physically sensible $\mathbf{P}$. Thus, fitting to data, and honing the precise details of the causal mechanisms and validating them becomes an exercise readily performed by the equations themselves. This is because the domain knowledge is high and the degrees of freedom are finite and low. 

Since the comprehensive theoretical scaffolding in field theory and a rich stationarity of action theory in physics is not present in the social sciences, less mathematical formalism is available to validate at a low degree of freedom to high precision. When equation systems corresponding to mechanistic changes are conjectured, refined and established as reliable knowledge, the equations themselves arise out of a combination of empirical observation and practical intuition, i.e., they are not deduced from an existing, deeper, formalism.

Consequently, a statistical model in Social Science could be a Generalized Linear Model, defined for a set of binary psychological traits as logits,
\[
p(A\vert B,C) = \left(1+\exp\left\{-\beta_0-\beta_1 B-\beta_2 C\right\}\right)^{-1}
\]
whose functional form exists for convenience. The system of history and genetic interactions is not encapsulated in any equation system explicitly. Nor can it, as the system of human behavior is irreducible. There is no mechanistic correspondence between this equation form and the actual governing processes of psychology and behavior. The model representation is very coarse, and thus any scientific empirical inquiry with it is incomplete, unlike solving inverse problems in physics as described above. 

Note the greater standards of statistical significance in the physical than social sciences, and lower out-of-distribution error. This is reflected in the contrasting replication rates of the fields. This is not to disparage social science, rather, it is important to indicate that the burden of proof lies not in the technical precision of physical instrumentation that is not accessible to the social sciences, but in establishing a preponderance of evidence by demonstrating a simultaneous consistent direction of causality by mechanisms at different layers of abstraction and scientific language games. Thus, the solution to the replication crisis in the social sciences is not to be found in adjustments in statistical practice, but rather concluding with less confidence towards suggesting that certain observations appear to be associated with each other. Or, with Bayesian model, the distribution of possible models within the model space depend on the empirical data distribution and any prior knowledge.

\paragraph{Causal Learning} We can observe that causal learning as it is defined concerns some statistical (e.g. usually linear structural equation model, but there are a number of nonlinear parametric, nonparametric, and neural variants) model but within a structure that resembles a forward physical model, i.e., for a multidimensional $X\in\mathbb{R}^{n_x}$,
\[
\dot{X}(t) = AX(t)+E(t)
\]
where $E(t)$ is a noise term and $A\in\mathbb{R}^{n_x\times n_x}$ is a matrix. Discovering causal relationships in the classical sense corresponds to the sparsity structure of $A$. However, it can be observed that 1) the assumption of causal sufficiency does not hold for most systems of interest, being open systems, 2) there is a significant bias in the choice of a statistical model here, as the mechanism is irreducible, and 3) in practice, finding sparsity in such models in the social sciences from real-world datasets is uncommon in the literature, and the few that do present type-1 error concerns.  

If, however, one is considering a special case of some (known to be exactly) linear dynamical system defining a dynamic process in an engineering device, then this model representation would be associated with a sparsity structure for $A$ defining the variable influences. In that case the sparsity structure properly identifies the causal relationship between quantities. As such the formalism, if it gets the equation form right, is correct for physical domains. 

On the other hand, consider that fitting a Dynamic Bayesian Network to time series data yields Figure~\ref{fig:DBN}, but there is still nontrivial variance in the model fit to the data, and out of distribution error. In that case, we can consider that $W_t$, compared to $R_t$ and $A_t$, is quite functionally distinct as far as unidirectionally influencing without being influenced. Otherwise, it appears that the direction of influence flows from $R_t$ to $A_t$. From this point, the appropriate next step is, rather than any conclusive declarations, proceed to, for example, performing physical experiments isolating $R_t$ and $A_t$, and finding the chemical composition of $W_t$. Science iterates, with knowledge and certainty being an asymptote that can always be made more rich in confirmed understanding.

The use of \textbf{do} calculus presents a case wherein a cause is conjectured, at least as far as an initial direct input. This suggests that it serves as a hypothesis test, as in a randomized trial. If the introduction of the \textbf{do} action, over many independent trials, is more likely to result in a certain outcome, then the scientist has proactively presented an interventionist causal hypothesis. Again,
at this point, investigation proceeds as based on the knowledge base and other known variables.

\paragraph{Neural Networks}

In circumstances wherein data samples are plentiful, modern developments in computing power and research and development in Machine Learning and AI has yielded Neural Networks as a transformative empirical tool. Through defining universal approximators in the form of sufficiently deep compositions of the appropriate nonlinear activation functions~\cite{scarselli1998universal}, Neural Networks can model any continuous functional relationship between quantities. Furthermore, through leveraging overparametrization and the theoretical properties of asymptotic regimes of NNs, one can see that they can simultaneously interpolate and generalize data (e.g.~\cite{shevchenko2022mean}). 

Still, the black box nature of NNs presents concerns as far as interpretability. In mission critical settings, for instance autonomous vehicles, hard safety should be constrained as far as, e.g. avoiding collisions, which cannot be guaranteed by a Deep Reinforcement Learning Model. As such, NNs must be considered as a particularly potent empirical model, but not one to infer the causal relationships in the universe by themselves. However, Recent developments in Scientific Machine Learning such as Physics Informed Neural Networks ~\cite{cuomo2022scientific} present a bridge whereby either physics-based structure, such as algebraic geometric domains, is enforced into the NN model, or physical requirements like conservation laws are maintained through some explicit physical formulation thereof. A promising development is their use in Drug Discovery, as due to molecular arrangement group structures.

One can see that in the case of a closed system experiment with a physics informed NN defining the model of the system, estimating, with sufficient data and parametrization, a NN that yields a sparsity structure between variables can imply likely causal relationships between variables. Here we see the alignment with the physics framework of causality. We observe that the tool of Scientifically Informed NNs is not available for the social sciences, presenting another asymmetry as far as causal modeling.

Finally, we remark on techniques to enhance NN understanding with mechanistic interpretability. We can see that here $\mathbf{h}_i$ and $\dot{\mathbf{y}}$ are latent variables in the statistical model. If we are able to identify $f(\mathbf{x})\approx\sum\limits_{i=1}^I f_i(\mathbf{x})$ and this produces similar outputs as $f(\mathbf{x})$ then we have identified a possible structure of mechanism, that is, sum-additive contributions through $I$ different pathways. One can then consider what class of object type in the system has cardinality $I$ and proceed with domain-level theoretical reasoning to attempt the next iteration of opening the black box. 

Likewise, the dimensionality of the autoencoder, that is $\vert\mathbf{W}\vert+\vert \mathbf{b}\vert$ that corresponds to good reconstruction loss, from which any lower dimensionality yields a steep drop-off, identifies a hidden degree of freedom associated with the mechanisms in the system. Thus such techniques \emph{interpret} the black box NN to yield topological properties of the possible set of \emph{mechanisms}.

\subsubsection{Mixed Methods Across Layers of Abstraction}
Consider now the study of a Complex Open System through the use of Mixed Methods Science Across Abstraction Layers, in order to establish a preponderance of evidence for causality in an open system. We consider both a biological and a social science domain of practice. 

First consider, from first principles, investigating physical activity's influence on a person's long term health and longevity. Different forms of statistical information would be arranged for - clinical trials of different activity mandates and biomarkers measured, detailed influence of physical activity on metabolic processes in the body, observational statistics gathered from survey data of exercise habits, past and present health metrics, survival analysis with hazard models depending on past and present exercise frequency, and others. Theory can appear as far as physical stimulus of sufficient intensity leading to physiological supercompensation, and thus greater integrity. Intermediary metrics, like the VO2Max, can be found to be a strong mediator, with an associated inference of lung health together with unrestricted and efficient oxygen flow along blood vessels throughout the body. Simulations can even be performed as far as cardiovascular fluid dynamics as far as developing a conjecture in silico of how exercise can be regenerative of the structural and functional integrity of the heart muscle. We can see the variety of both mechanical simulations that demonstrate how the cause is manifested to assist in completing the picture that empirical observations had drawn with a less sharply defined and more ambiguous imagery. 

Consider next a circumstance wherein a human is interacting with other humans in an environment defined by institutions. The interaction is both physical and social, and a number of statistical observations of social behavior are taken. For instance, suppose one were to inquire as to the potential influence of certain childhood experiences on behavior in professional settings as an adult. Statistical observations can arise in the form of surveys of experiences of childhood and professional social engagement and twin cohort observation studies can be used to modulate genetic influence. In addition, clinical trials of finding the significance and effect size of therapeutic intervention in the intention of removing the influence of these experiences on current cognition and behavior can be performed. This statistical work can be accompanied by Hermeneutic work, for instance finding common meta-narratives such as ``there is great risk in even the slightest error'' in the two settings. Psychological theory can inform through, e.g. Jungian archetype analysis of the narrative and therapeutic practice of assuaging traumatic influence and personality traits in the five factor model that mediate how the experience influences adult behavior are both Theory-based scientific research towards this topic. Finally, an explanation of the evolutionary mechanism can be developed through the use of early life environment signaling models: a childhood experience was in prehistoric times more representative of the world around it and thus informative as to its most salient properties of safety and trust, so observing chaos in childhood appropriately causes greater vigilance in adulthood (which may be maladaptive in the modern civilized world). A conclusive finding would be simultaneous harmony across all of the statistical domains together with sensible theoretical narratives.

Finally, we should note that epistemic virtue requires forsaking one's education-identified elitism and considering folk knowledge. Through experience and practice, practitioners, or clinicians, in a field may stumble onto a truth before formal science recognizes its validity. This has coincided with formal establishment science disparaging the validity of a practice before later confirming it with formal methods. This has occurred historically, for instance, as far as the use of meditation and psychedelics for mental health, for which clinical trials showing positive efficacy have only recently been consistently appearing in the scientific literature. Thus, it is recommended that any Mixed Methods Scientist also considers the possibility of clinicians in the field, both market professionals and communities of amateurs, having information to cross compare with the findings in the other domains. If they must interact with the system for some useful aim, they are incentivized to do it with as accurate 
world model of the system as possible. Purpose-driven thought is another cognitive framework, whose application and corrections to real-time data can yield insights into a system's patterns and structure of behavior. In health and medical domains, for example, general practitioners who may see patterns across patients with respect to activity level and frequency of illness, athletic coaches who can see the effect of various exercise regimes, support staff at retirement communities witnessing potentially undiscovered side effects of medication, etc.

We remark that an important aspect of scientific practice under these principles is the translation of ideas and processes across disciplines. Given the presence of completely distinct paradigms and means of understanding the domain, this requires significant attention to principled interdisciplinary research.  One can observe that the example of cognitive science provided multiple examples of integration schemas. As a potential direction to investigate, observe that analogies are central to human cognition as far as leaping across domains through abstract reasoning~\cite{mitchell2001analogy}. Moreover, analogies can be mapped to first-order logic~\cite{abdelfattah2019semantics}. We then see means of performing a concatenation of reasoning by formalizing domains in CP-Logic and performing analogous reasoning. 
For example, we know that cold temperature (\texttt{Cold}) generally causes living organisms (\texttt{Living}) to more likely ($\alpha$) shiver: 
$
(\texttt{Shiver($x$)} : \alpha) \leftarrow \texttt{Cold($x$)} \wedge \texttt{Living($x$)}.
$
By analogy ($\texttt{Analogous}(R_1, R_2, f)$), i.e. some structure-preserving mapping $f$ across domains: $\forall x, y \quad R_1(x, y) \rightarrow R_2(f(x), f(y))$, we may recognize that cold temperature corresponds to slow-moving molecules ($R_1(\texttt{Cold},\texttt{SlowM})$) as shivering corresponds to input of kinetic energy ($R_2(\texttt{Shiver},\texttt{KineticE})$). Consequently, we may infer that slow moving molecules tend to receive injections of kinetic energy 
$
(\texttt{KineticE($x$)} : \alpha) \leftarrow \texttt{SlowM($x$)} \wedge \texttt{Living($x$)}
$
in systems where living organisms attempt to maintain homeostasis.


\subsection{Final Thoughts on Scientific Mindset}

We hope that this work provides sufficient force of argument for exhibiting greater intellectual humility in the sciences. At the same time, we hope that rather than engendering disappointment and demotivating scientific enterprise, it recognizes that the challenge of epistemic virtue in expressing degrees of certainty provides a rich Research Program in and of itself. Practice becomes less about debate in conflict, but harmonization and integration. Rather than weighing the relative validity of alternative explanations, a research program that prioritizes reconciling them and developing best practices in interdisciplinary research provides for a boundless set of problems of heterogeneous intellectual flavor. Each such flavor enriches the palate of understanding and comprehension, for which perspectives across domains and levels of abstraction have a number of routes for the establishment of harmony, rather than rivaling sports teams vying for hegemony. 

\section*{Acknowledgments}
The first author would like to acknowledge Tim Snow and Joyce Carpenter. About twenty years ago, the first author, in his youth, met Tim and Joyce, in their young adulthood a few years after their Philosophy PhD studies, on an internet discussion forum dedicated to intellectual discussions. Tim and Joyce graciously proceeded to tutor VK through a comprehensive selection of prominent works in Western Epistemology from Descartes to Quine. A few years ago when the idea for this paper came to VK, he re-established contact with them after two decades without, and while they were not able to commit time to help with working on the paper, they provided valuable discussion and encouragement. As such, this work would not have been written without them, for multiple reasons. 

\bibliographystyle{plain}
\bibliography{refs.bib}
\end{document}